\documentclass[lettersize,journal]{IEEEtran}

\IEEEoverridecommandlockouts

\usepackage{cite}
\usepackage{amsmath,amssymb,amsfonts}
\usepackage{graphics}
\usepackage{graphicx}
\usepackage{textcomp}
\usepackage{xcolor}
\usepackage{subfigure} 
\makeatletter

\usepackage{array}
\usepackage{multirow}
\usepackage{longtable}
\usepackage{rotating}
\usepackage{booktabs}

\usepackage{cite}
\usepackage{amsmath,amssymb,amsfonts}
\usepackage{arydshln}
\usepackage{graphics}
\usepackage{graphicx}
\usepackage{textcomp}
\usepackage{xcolor}
\usepackage{subfigure} 
\makeatletter  

\newcommand{\Rmnum}[1]{\expandafter\@slowromancap\romannumeral #1@}
\newif\if@restonecol  
\makeatother

\usepackage[linesnumbered,ruled,vlined]{algorithm2e}
\usepackage{algpseudocode}  
\usepackage{amsmath}  
\newtheorem{assumption}{Assumption}

\usepackage{array}
\usepackage{multirow}
\usepackage{longtable}
\usepackage{rotating}
\usepackage{booktabs}

\usepackage{amsmath,amsfonts}

\usepackage[caption=false,font=normalsize,labelfont=sf,textfont=sf]{subfig}
\usepackage{textcomp}
\usepackage{stfloats}
\usepackage{xurl}
\usepackage{verbatim}
\usepackage{graphicx}
\usepackage{cite}
\begin{document}

\title{Large-Scale OD Matrix Estimation with A Deep Learning Method}

\author{IEEE Publication Technology,~\IEEEmembership{Staff,~IEEE,}
\author{\IEEEauthorblockN{Zheli Xiong\thanks{Z. Xiong is with the School of Data Science, University of Science and Technology of China, Hefei, Anhui 230000, China. E-mail: zlxiong@mail.ustc.edu.cn.}, Defu Lian{*}\thanks{D. Lian and E. Chen are with the Anhui Province Key Laboratory of Big Data Analysis and Application, School of Computer Science and Technology, University of Science and Technology of China, Hefei, Anhui 230000, China. E-mail: \{liandefu, cheneh\} @ustc.edu.cn.}, Enhong Chen, Gang Chen\thanks{G. Chen and X. Cheng are with the Yangtze River Delta Information Intelligence Innovation Research Institute, China. Email: cheng@ustc.win, wh5606@ustc.edu.cn} and Xiaomin Cheng\thanks{Corresponding author: Defu Lian}}\\ }

\thanks{This paper was produced by the IEEE Publication Technology Group. They are in Piscataway, NJ.}
\thanks{Manuscript received April 19, 2021; revised August 16, 2021.}}

\markboth{Journal of \LaTeX\ Class Files,~Vol.~14, No.~8, August~2021}%
{Shell \MakeLowercase{\textit{et al.}}: A Sample Article Using IEEEtran.cls for IEEE Journals}


\maketitle

\begin{abstract}
The estimation of origin-destination (OD) matrices is a crucial aspect of Intelligent Transport Systems (ITS). It involves adjusting an initial OD matrix by regressing the current observations like traffic counts of road sections (e.g., using least squares). However, the OD estimation problem lacks sufficient constraints and is mathematically underdetermined. To alleviate this problem, some researchers incorporate a prior OD matrix as a target in the regression to provide more structural constraints. However, this approach is highly dependent on the existing prior matrix, which may be outdated. Others add structural constraints through sensor data, such as vehicle trajectory and speed, which can reflect more current structural constraints in real-time.
Our proposed method integrates deep learning and numerical optimization algorithms to infer matrix structure and guide numerical optimization. This approach combines the advantages of both deep learning and numerical optimization algorithms. The neural network(NN) learns to infer structural constraints from probe traffic flows, eliminating dependence on prior information and providing real-time performance. Additionally, due to the generalization capability of NN, this method is economical in engineering. We conducted tests to demonstrate the good generalization performance of our method on a large-scale synthetic dataset. Subsequently, we verified the stability of our method on real traffic data. Our experiments provided confirmation of the benefits of combining NN and numerical optimization.
\end{abstract}


\section{Introduction}
With the advent of traffic big data, data mining has enabled the application of various traffic tasks, among which prediction and optimal control tasks are of utmost importance. Prediction tasks involve forecasting future traffic conditions and behaviors based on historical traffic data, such as traffic flow prediction \cite{li2017diffusion}, trajectory prediction \cite{altche2017lstm} , and location recommendation \cite{lian2020geography}. In control tasks, such as traffic light control on Intelligent Transport Systems (ITS), traffic data is utilized to support decision-making. Traffic demand plays a crucial role in the aforementioned tasks. Firstly, different traffic demands can impact the next period's traffic condition and routing decision or recommendation. Secondly, traffic demand can provide better insights into ITS decisions. However, unlike ordinary traffic data, traffic demand is not directly collected but can be obtained through OD estimation, which aims to utilize the surveyed prior OD and traffic counts to estimate the traffic demand from each origin to every other destination on the transport network in a period.

OD estimation is primarily employed in static OD estimation \cite{behara2020novel} and offline cases in dynamic OD situations \cite{cascetta1993dynamic}. It involves adjusting an initial OD matrix through regression on the observable traffic counts of road sections (e.g., least squares). In contrast to OD prediction, which utilizes the evolution of historical OD sequences to predict the OD matrix in the next period, OD estimation considers that historical OD sequences are not available and relies solely on inferring the OD matrix from the current observation data. OD estimation is a useful technique for generating an initial historical OD matrix that can be used for OD prediction purposes. The OD estimation problem lacks sufficient constraints and is mathematically underdetermined \cite{robillard1975estimating}. Therefore, the key is to extract more constraints from prior knowledge of the OD matrix and observable data.

One common approach to solving OD estimation problems is the bi-level framework \cite{bert2009dynamic}, which consists of upper and lower levels. The upper level is tasked with adjusting the OD matrix by minimizing the numerical gap between real and estimated traffic counts through the construction of a convex optimization problem (Eq. 1). The optimization method used in the upper level is predominantly gradient-based \cite{spiess1990gradient}, specifically the steepest descent method. Meanwhile, the lower level assigns demand to road sections through analysis \cite{maher2001bi} or simulation \cite{behara2020novel} methods. By leveraging fixed point theory, the upper and lower levels update iteratively and converge to an optimal solution.

To address the problem of underdetermination, several papers have proposed different methods for adding constraints, such as travel time \cite{barcelo2010travel}, travel speed \cite{jaume2015integrated}, and turning proportions \cite{alibabai2008dynamic}. Some researchers use a prior surveyed target OD matrix \cite{cascetta1984estimation}\cite{lundgren2008heuristic} or prior sampled OD matrix to maintain consistency in the structure of the OD matrix. Alternatively, structure information can be introduced through vehicle trajectory data, such as cellular probe \cite{calabrese2011estimating} or Bluetooth probe \cite{behara2020novel}, to alleviate the underdetermined problem and help converge towards a better solution point to avoid falling into a local optimum prematurely.

To our knowledge, no previous study has proposed using deep learning to solve the OD estimation problem. Due to its powerful fitting ability, the NN can learn the mapping relationship from traffic counts to the OD matrix. However, the very large numerical sampling space can cause the real data to become an outlier outside the training data, as the unrestricted values range from 0 to large number. To overcome this issue, we use restricted mappings such as distributions ranging from 0 to 1, which limit the space's size and reflect the structural constraints. Further details are provided in Chapter 4. Additionally, numerical predictions using deep learning may become inaccurate due to the lack of assignment information from the OD matrix to traffic counts. Therefore, it is necessary to incorporate numerical optimization methods to perform assignment calculations. We combine the advantages of both deep learning and numerical optimization algorithms. The NN learns to infer structural constraints from probe traffic flows, due to the generalization capability of NNs, it is an economical solution in engineering.

In this paper, we further deliver the following contributions.
\begin{itemize}
\item The effectiveness and generalization performance of our proposed method are demonstrated through evaluation on both a large-scale synthetic dataset and a real traffic dataset.
\item We have demonstrated through experiments that this deep-learning framework has the capability to integrate various advanced estimation methods, thereby achieving better performance. This indicates that our framework possesses scalability.
\item The proposed method provides a valuable solution with respect to accuracy, stability, and economy.

\end{itemize}

\noindent\textbf{Reproducibility.} The code and reproduction package are available at \url{https://github.com/shaun19920309/Large-Scale-OD-Matrix-Estimation-with-A-Deep-Learning-Method}.

\section{Related Works}
\subsection{OD matrices estimation}\label{AA}
The fundamental goal of OD estimation is to leverage prior knowledge of OD-flow and observed variables to construct an optimization function and establish the relationship between OD-flow and traffic counts using the assignment model.

\subsubsection{Optimization}\label{AA}

Several approaches have been proposed to estimate OD matrices through optimization of the objective function using prior knowledge of OD flow and observed variables. The gravity model, introduced by Lowry \cite{low1972new} and Högberg \cite{hogberg1976estimation}, assumes a "gravitational behavior" for travel demand and constructs a linear or nonlinear regression model. The maximum likelihood method, proposed by Spiess \cite{spiess1987maximum}, assumes that OD pairs follow independent Poisson distributions and estimates the OD matrix by maximizing the likelihood over the sampled matrix. This method can be viewed as optimization based on distribution since maximum likelihood is equivalent to optimizing KLD (Kullback–Leibler divergence). The entropy maximizing/information minimization method, introduced by Van Zuylen \cite{van1980most}, aims to choose an estimated OD matrix that adds as little information as possible contained in traffic counts to address the underdetermined problem. The Bayesian approach \cite{maher1983inferences} estimates OD matrices by maximizing the posterior probability, which is a combination of the prior OD matrix and observations. The maximizing/information minimization method is a specific case of this approach when there is minimal confidence in the prior. Generalized least squares (GLS) \cite{cascetta1984estimation} explicitly considers biases in observed flow and the target OD matrix. The previously described methods necessitate the utilization of prior OD matrices for parameter estimation. These matrices can be acquired through surveys and other means, which can be costly and time-consuming. Moreover, the matrices may be outdated and inaccurate, leading to significant errors. With the availability of big traffic data and traffic simulation, other variables such as travel time \cite{barcelo2010travel}, travel speed \cite{jaume2015integrated}, and turning proportions \cite{alibabai2008dynamic} can also be used directly in OD estimation.

Directly estimating the OD matrix by introducing various types of data can be challenging due to the difficulty of effectively fusing the data and ensuring extendibility. To address this issue, we have implemented a deep learning model that can fuse multivariate data to accurately infer the matrix structure. Additionally, we have incorporated our proposed fusion optimization method into the deep learning model to create a extensible OD estimation framework.


\subsubsection{Assignment}\label{AA}

Assignment models are commonly used to allocate traffic flows based on shortest paths, such as Moore (1957) and Dijkstra (1959) algorithms. Route choice is a key factor in determining the allocation of vehicles on these paths. The most basic form of assignment model is the static assignment model, which assumes that the cost of each link does not change. The all-or-nothing (0-1) assignment model is an example of a static assignment model, which assigns a value of 1 if the least cost path passes through a specific link and 0 otherwise. On the other hand, dynamic assignment models consider the current network situation when allocating the least cost path, such as the Stochastic Route Choice (SRC) model, which uses historical performance to determine the current network situation. Dynamic assignment modes, such as the Logit and C-logit models, have also been proposed. However, since the path is determined based on the current situation, it does not guarantee traffic equilibrium within the network. To address this issue, researchers have proposed equilibrium models. Wardrop equilibrium is a static equilibrium model that follows the User Equilibrium (UE) principle. It states that "Under equilibrium conditions, traffic arranges itself in congested networks in such a way that no individual trip maker can reduce his path cost by switching routes." This corresponds to the second principle "Under social equilibrium conditions traffic should be arranged in congested networks in such a way that the average (or total) travel cost is minimized." The static equilibrium model cannot maintain network equilibrium in the dynamic process when traffic conditions are not stable. Therefore, the Dynamic User Equilibrium (DUE) model has been proposed as a dynamic version of Wardrop's user equilibrium, which is more consistent with reality.

\subsection{OD matrices prediction}\label{AA}

\subsubsection{Traditional method}\label{AA}

OD prediction and OD estimation differ in their applicability to dynamic OD situations, with OD prediction being more suitable for online scenarios. In OD prediction, the future OD matrix is predicted using the evolution law inferred from the historical OD sequence. The main methods used for OD prediction include the Kalman filter (KF) \cite{bierlaire2004efficient}, Constrained Kalman Filter (CKF) \cite{simon2005aircraft}, and LSQR \cite{paige1982lsqr}.

\subsubsection{Deep learning method}\label{AA}

The ability to extract and fuse features makes NNs a promising approach for dynamic OD matrix prediction. To capture spatial features, researchers use methods such as Graph Neural Networks (GNN) \cite{chen2017supervised} or Graph Convolutional Networks (GCN) \cite{xiong2020dynamic} on the city graph. Additionally, LSTM (Long Short-Term Memory) \cite{chu2019deep} or GRU (Gated Recurrent Unit) \cite{pan2019urban} methods are used to capture the historical changes in traffic trips.

\section{Preliminary}
\subsection{Definition}\label{AA}
In the city network, an OD node is a cluster obtained by clustering the intersections of road sections. The set of OD nodes is denoted as $\pmb{\upsilon}=\{\pmb{n}_{1},\pmb{n}_{2},...,\pmb{n}_{|\pmb{\upsilon}|}\}$, and there are $|\pmb{\upsilon}|$ OD nodes in the study network. The traffic counts of aggregated road sections are denoted as $\pmb{\epsilon}=\{e_{1},e_{2},...,e_{|\pmb{\epsilon}|}\}$, where $|\pmb{\epsilon}|$ is the number of aggregated road sections. These sections aggregate road sections that connect the same OD pair, as shown in Fig. 3. The OD matrix, $\mathbf{M}$, represents the traffic flow between each OD node, where $\mathbf{M}_{ij}$ represents the traffic flow from OD node $\pmb{n}_{i}$ to OD node $\pmb{n}_{j}$.

Each OD node $\pmb{n}_{i}\in \mathbb{R}^{|\pmb{\epsilon}|}$ is represented by a vector of length $|\pmb{\epsilon}|$, where $\pmb{n}_{ij}=1$ if road section $j$ enters node $\pmb{n}_{i}$, $\pmb{n}_{ij}=-1$ if it exits from $\pmb{n}_{i}$, and $\pmb{n}_{ij}=0$ if it is not connected to $\pmb{n}_{i}$. This allows for the representation of the topology of the road network. The traffic count distribution $\pmb{d}_{\epsilon}$ is obtained by normalizing the traffic counts, $\pmb{d}_{{\epsilon}{j}}=\frac{\pmb{\epsilon}_{j}}{\sum\limits_{i=1}^{|\pmb{\epsilon}|} \pmb{\epsilon}_{i}}$.

The production flow represents the total trips departing from each node, $\pmb{p}=\sum\limits_{j=1}^{|\pmb{\upsilon}|}\mathbf{M}_{ij}$. The attraction traffic represents the total trips arriving at each node, $\pmb{a}=\sum\limits_{i}^{|\pmb{\upsilon}|}\mathbf{M}_{ij}$. The production flow distribution $\pmb{d}_{p}$ is obtained by normalizing the production flow, $\pmb{d}_{pj}=\frac{\pmb{p}_{j}}{\sum\limits_{i=1}^{|\pmb{\upsilon}|} \pmb{p}_{i}}$, and the attraction flow distribution $\pmb{d}_{a}$ is obtained by normalizing the attraction flow, $\pmb{d}_{aj}=\frac{\pmb{a}_{j}}{\sum\limits_{i=1}^{|\pmb{\upsilon}|} \pmb{a}_{i}}$. $\pmb{d}_{p}^{*}$ represents the best-inferred distribution of $\pmb{p}$, and $\pmb{d}_{a}^{*}$ represents the best-inferred distribution of $\pmb{a}$.

In the optimization process, $\hat{\pmb{d}}_{p}$ and $\hat{\pmb{d}}_{a}$ correspond to $\hat{\pmb{p}}$ and $\hat{\pmb{a}}$, respectively, in each iteration.

\subsection{Numerical optimization}\label{AA}
Flattening the OD matrix to a vector makes it easier to represent and manipulate the traffic demand between OD pairs. The vector $\pmb{t}$ has length $|\pmb{\upsilon}|^{2}$ because there are $|\pmb{\upsilon}|$ OD nodes, and the traffic demand between each pair of OD nodes is represented by a single element in the vector. In the optimization process, the goal is to estimate the actual traffic demand in each OD pair, which is represented by the vector $\hat{\pmb{t}}$. Thus,
the traditional optimization function is formulated as below:
\begin{IEEEeqnarray}{c} 
\min_{\hat{\pmb{t}}}Z(\hat{\pmb{t}})=\min_{\hat{\pmb{t}}}\frac{1}{2}(\pmb{\epsilon}-\hat{\pmb{\epsilon}})^{\mathrm{T}}(\pmb{\epsilon}-\hat{\pmb{\epsilon}}) \IEEEnonumber\\  
where\  \hat{\pmb{\epsilon}}=\pmb{P}\hat{\pmb{t}}
\end{IEEEeqnarray}
The assignment matrix, denoted by $\pmb{P} \in \mathbb{R}^{|\pmb{\epsilon}| \times |\pmb{\upsilon}|^{2}}$, represents the assignment ratio of each OD trip to each road section. The matrix $\pmb{P}$ can be obtained through an analytical method, such as the stochastic user equilibrium approach based on traffic counts proposed in \cite{maher2001bi}, or by simulation using a traffic simulator like SUMO \cite{lopez2018microscopic}. In our study, we estimate the assignment matrix $\pmb{P}$ using a back-calculation procedure based on simulated traffic counts obtained from the traffic simulator in each iteration, following a similar approach to \cite{behara2020novel}.

After each iteration, using gradient decent method to update $\hat{\pmb{t}}$:
\begin{equation}
\hat{\pmb{t}}^{k+1}=\hat{\pmb{t}}^{k} \odot (\pmb{e}- \lambda^{k} \odot \nabla Z(\hat{\pmb{t}}^k)
\end{equation}
where $\pmb{e}$ is a vector of 1s and of dimension same as $\hat{\pmb{t}}$, $\odot$ denotes element wise product. The second optimization determines the best step length at $k$'th iteration with non-negative constraint:
\begin{IEEEeqnarray}{c} 
\min_{\lambda^{k}}Z((\hat{\pmb{t}}^{k}_{i}(1-\lambda^{k}   \nabla Z(\hat{\pmb{t}}^{k}_{i}))) \IEEEnonumber\\
s.t.\  \lambda^{k}  \nabla Z(\hat{\pmb{t}}^{k}_{i}) \leq 1\\
for \ all\ i \in |\pmb{\upsilon}|^{2} \ with\  \hat{\pmb{t}}^{k}_{i}>0 \IEEEnonumber
\end{IEEEeqnarray}

\subsection{Distribution similarity measure}\label{AA}
The Kullback-Leibler Divergence (KLD) is a measure of the similarity between two probability distributions: the real distribution $P$ and an approximate distribution $Q$. The Forward KLD, denoted $KL_{f}(P||Q)$, ensures that wherever $P$ has a high probability, $Q$ must also have a high probability. This property tends to avoid any position where $P$ is greater than 0, but $Q$ is equal to 0, which is called zero avoiding. This behaviour is akin to Mean-Seeking. On the other hand, the Reverse KLD, denoted $KL_{r}(Q||P)$, guarantees that wherever $Q$ has a high probability, $P$ must also have a high probability. This property ensures that $Q$ contains the higher probability part of $P$, which is called Mode-Seeking.

However, KLD is not symmetric and cannot be used to measure how close two distributions are. In contrast, JSD (Jensen-Shannon Divergence) can symmetrically measure the similarity of two distributions. The two formulas are as follows.
\begin{equation}
KL(P||Q)=\sum\limits_{i=1}^{n}P_{i}\log \frac{P_{i}}{Q_{i}}
\end{equation}
\begin{equation}
JS(P||Q)=\frac{1}{2}KL(P||\frac{P+Q}{2})+\frac{1}{2}KL(Q||\frac{P+Q}{2})
\end{equation}

\section{Method}

In this chapter, we will provide a detailed explanation of the pipeline of our proposed method. This includes sampling probe traffic to obtain training datasets, using a NN model, and integrating the production flow distribution $\pmb{d}_{p}$ and attraction flow distribution $\pmb{d}_{a}$ obtained from deep learning into numerical optimization.

\subsection{Probe Traffic Sampling}\label{AA}
In order for the Distribution Learner to effectively learn the mapping relationship between traffic counts distribution and OD distribution, the training data must be carefully selected. The samples used for training should be distributed around the real samples, in order to prevent the true distribution from becoming an outlier during training. To achieve this, we first make an following assumption about the OD distribution in a real city.

\begin{assumption}
The production and attraction flows of urban origin-destination pairs are not evenly distributed. In other words, there are variations in the total number of departures and arrivals from different locations. This unevenness applies to both the production flow $\pmb{p}$ and attraction flow $\pmb{a}$.
\end{assumption}

This assumption is practical because it acknowledges that not all nodes have the same number of vehicles departing and arriving. Uneven distributions can provide more valuable structural information than uniform distributions because uniform distributions indicate a lack of information to extract. To create such a non-uniform distribution, we sample the sparse OD matrix with a density parameter $\varepsilon$.

Looking at the urban OD matrix from another perspective, the flows that truly make $\pmb{p}$ and $\pmb{a}$ uneven are concentrated in a few critical OD pairs, while most of the other flows are relatively small (as shown in Fig. 1(a)). Therefore, the matrix can be approximated as a sparse matrix. The probe vehicle flows will appear in these critical OD pairs, enabling the detection of structural information.

To summarize, the assumption is that the production and attraction flows of urban OD are uneven, and we sample the sparse OD matrix with a density parameter to create a non-uniform distribution. We randomly sample non-zero OD pairs to represent the critical OD pairs and use uniform sampling with a maximum value of M for the non-zero OD pairs. We compare the resulting $\pmb{d}_{p}$ and $\pmb{d}_{a}$ with the uniform distribution $St=\frac{\pmb{e}}{|\pmb{\upsilon}|}$, where $\pmb{e}$ is a vector of 1s, to ensure that they are non-uniform. The degree of non-uniformity is represented by $\beta$, and we require that $KL_{r}(\pmb{d}_{p}||St)>\beta$ and $KL{r}(\pmb{d}_{a}||St)>\beta$ for the sampled matrix to be considered suitable. Our training data consists of pairs $(\pmb{d}_{p},\pmb{d}_{a})$, with the generated OD matrix $\mathbf{M}$ as the label and the corresponding $\pmb{d}_{\epsilon}$ of simulated traffic counts vector $\pmb{\epsilon}$ as the input (see Fig. 1(b)).

\begin{figure}[htbp]
\subfigure[Urben OD matrix]{\includegraphics[width=4cm]{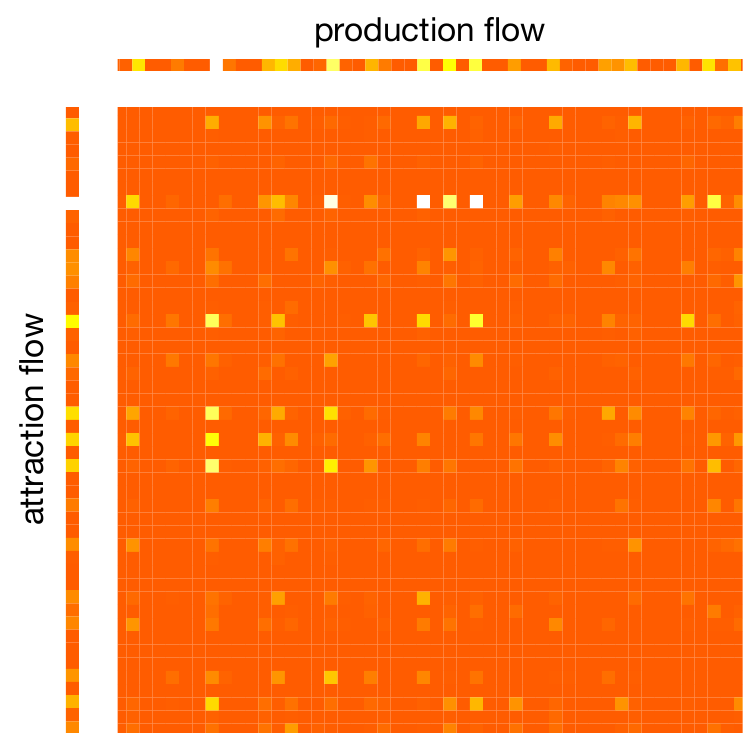}}
\subfigure[Sampling matrices]{\includegraphics[width=4.3cm,height=3.5cm]{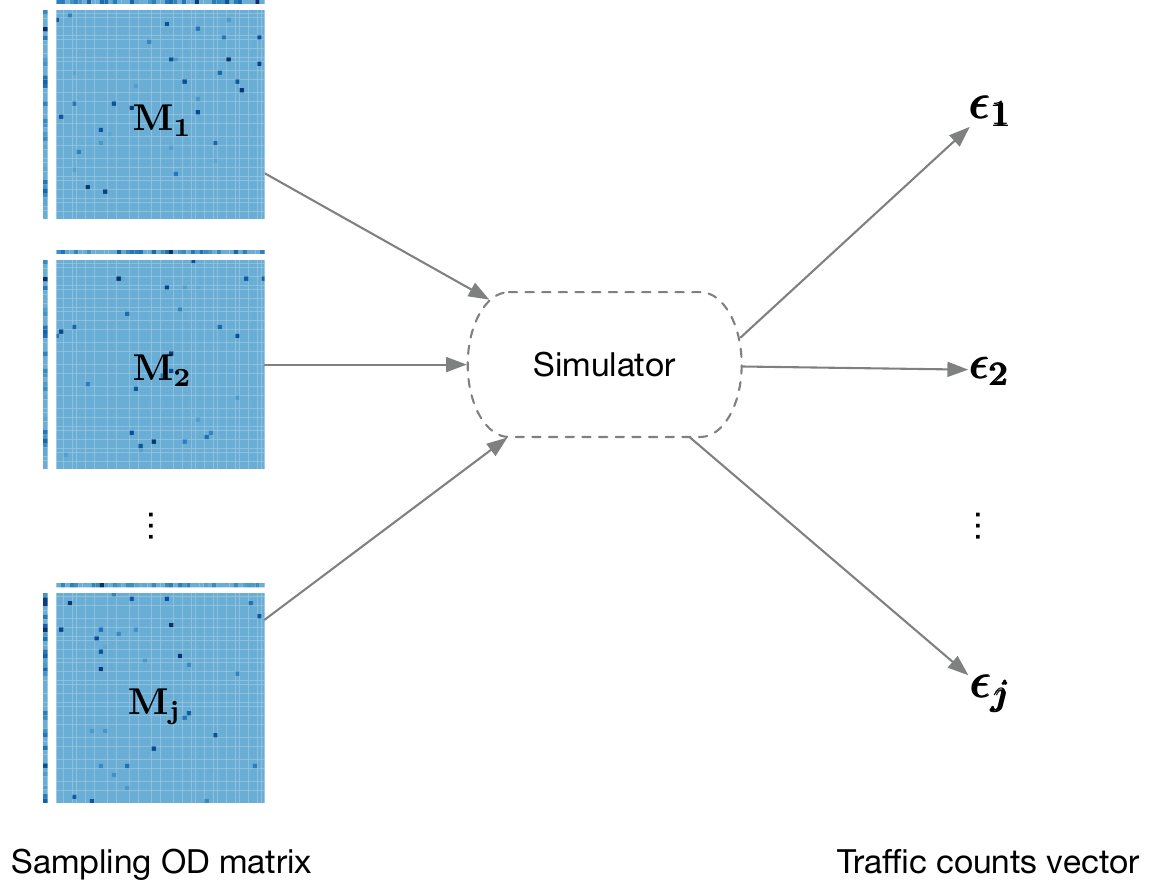}}
\caption{(a) Urben OD matrix, production flow and attraction flow. Highlighting indicates heavy traffic flow. (b) Sampling matrices}
\label{fig}
\end{figure}

\subsection{Distribution Inference}\label{AA}
In the distribution inference process, Our model play a function $F$, which takes topology information of nodes $\pmb{\upsilon}$ and traffic counts distribution $ \pmb{d}_{\epsilon}$ as inputs and output an inferred production distribution $\bar{\pmb{d}}_{p}$ or attraction distribution $\bar{\pmb{d}}_{a}$.

\begin{equation}
\bar{\pmb{d}}_{p\ or\ a}=F_{p\ or\ a}(\pmb{\upsilon}, \pmb{d}_{\epsilon})
\end{equation}

The interactions between each pair of OD nodes cannot be ignored. For example, suppose two adjacent OD nodes, $n_{i}$ and $n_{j}$, have traffic trips passing through the same road section, $e$. In that case, if the traffic trip of $n_{i}$ increases, the traffic trip of $n_{j}$ should decrease, assuming the traffic counts of $e$ are fixed. Additionally, the sum of the elements in the OD distribution must equal 1, and every element is constrained between 0 and 1. Hence, we have chosen to employ the Transformer technology as our network architecture, which was inspired by recent researchers\cite{dosovitskiy2020image} and \cite{he2021masked}. They proposed a Transformer NN to address a Computer Vision (CV) task similar to our demand, where an image with a large-scale invisible masked pixel was given as input, and the masked pixel value needed to be inferred to recover the entire image. The relevance of this task to our work is reflected in two aspects: a) unknown elements must be inferred by considering known elements and their dependencies, and b) inferred elements must be restricted (e.g., image pixel values are limited from 0 to 255).

The Transformer frame employs the Multihead Self-Attention (MSA) module and an efficient structure\cite{vaswani2017attention}. The upstream encoder captures pairwise dependencies between two elements and represents the sequence input, while the decoder downstream is used to perform inference based on various scenarios. The Transformer frame has been widely used in Natural Language Processing (NLP) tasks \cite{devlin2018bert} such as machine translation \cite{wang2019learning} in the past due to its strong capability of capturing the grammatical correlation between words in any long distance in a sentence.

\begin{figure*}[htbp]
\centerline{\includegraphics[width=15cm,height=5cm]{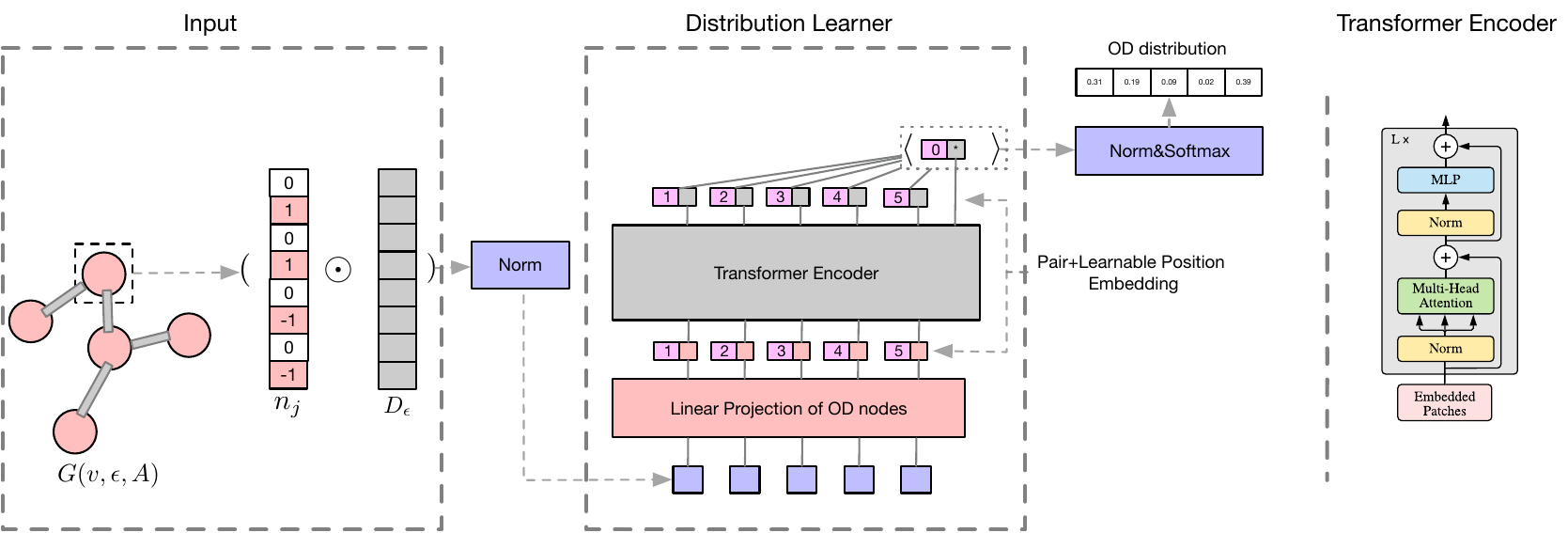}}
\caption{Distribution Learner}
\label{fig}
\end{figure*}

\subsubsection{Multihead Self-Attention(MSA)}
Standard $\pmb{qkv}$ self-attention(SA) \cite{vaswani2017attention} compute the attention weight $A$ over all value of elements $\pmb{v}$. $A$ is based on the of query $\pmb{q}$ and key $\pmb{k}$ of elements, and calculate the pairwise dependency between two elements of input sequence $\pmb{\zeta} \in \mathbb{R}^{(|\pmb{\upsilon}|^{2}+1) \times d}$.
\begin{equation}
[\pmb{q},\pmb{k},\pmb{v}]=\pmb{\zeta}\pmb{U}_{qkv},\  \pmb{U}_{qkv} \in \mathbb{R}^{d \times 3d_{h}}
\end{equation}
\begin{equation}
A=softmax(\pmb{qk}^\mathrm{T} / \sqrt{d_{h}})
\end{equation}
\begin{equation}
SA(\pmb{\zeta})=A\pmb{v}
\end{equation}

MSA executes $h$ SA operations in parallel and projects the concatenated outputs. $h$ is the number of heads, in order to keep the dimensions unchanged, $d_{h}$ is set to $d/h$.
\begin{IEEEeqnarray}{c} 
MSA(\pmb{\zeta})=[SA_{1}(\pmb{\zeta});SA_{2}(\pmb{\zeta});...;SA_{h}(\pmb{\zeta})]\pmb{U}_{msa}, 
\IEEEnonumber\\
\pmb{U}_{msa} \in \mathbb{R}^{h \cdot d_{h} \times d} 
\end{IEEEeqnarray}
 $\pmb{U}_{qkv}$  and $\pmb{U}_{msa}$ above are learnable parameters

\subsubsection{Distribution Learner}

In the following section, we will discuss the composition of Distribution Learner and the different components involved.

To begin with, we combine topology information and road sections distribution information by utilizing element-wise multiplication of $\pmb{\upsilon}$ and $\pmb{d}_{\epsilon}$, which is then used as input for the model.

\begin{equation}
\pmb{\chi}=\pmb{\upsilon} \odot \pmb{d}_{\epsilon}
\end{equation}

The overall structure of the Distribution Learner is presented in Fig. 2. Initially, we pass the input into a trainable linear projection and refer to the output of projection as node embedding. We incorporate standard learnable $1d$ position embedding to the node embedding to retain the positional information. Subsequently, the resultant embedding vector is passed through L layers of a standard Transformer encoder with an alternating structure.

The structure of each encoder layer is consistent, with the output of the last layer encoder serving as input to the subsequent layer encoder. The encoder structure, as illustrated in Fig. 2, comprises alternating layers of MSA and MLP blocks. Before every block, Layer Norm (LN) is applied, and residual connections are implemented after every block, as suggested in \cite{wang2019learning} and \cite{baevski2018adaptive}. The MLP consists of two layers, and a $tanh$ activation function is incorporated in each layer, following the approach of \cite{dosovitskiy2020image}.

To recover the position information and obtain the final node representations, position embedding is added once again to the output embedding vectors of the L layers Transformer encoder. As the distribution is node-related, we aggregate all the final representations to obtain a global representation vector, denoted as $cls$, which serves a similar purpose to BERT’s $[class]$ token. We then compute the inner-product between each node representation and $[class]$ token to obtain a scalar value for each node. Finally, we perform a Layer Norm (LN) and apply the softmax operation to generate the final output inference in the form of OD distribution, denoted as $\bar{\pmb{d}}$.

We formulated the whole process as below: 
\begin{IEEEeqnarray}{c} 
\pmb{\zeta}_{0}=[\pmb{\chi}_{1}\pmb{E};\pmb{\chi}_{2}\pmb{E};...;\pmb{\chi}_{|\pmb{\upsilon}|}\pmb{E}]+\pmb{E}_{pos},\IEEEnonumber\\
\pmb{E} \in \mathbb{R}^{|\pmb{\epsilon}| \times d},\pmb{E}_{pos} \in \mathbb{R}^{|\pmb{\upsilon}| \times d}
\end{IEEEeqnarray}

\begin{equation}
\pmb{\zeta'}_{\iota}=MSA(LN(\pmb{\zeta}_{\iota-1}))+\pmb{\zeta}_{\iota-1},\ \iota=1...L 
\end{equation}
\begin{equation}
\pmb{\zeta}_{\iota}=MLP(LN(\pmb{\zeta'}_{\iota}))+\pmb{\zeta'}_{\iota},\ \iota=1...L 
\end{equation}
\begin{equation}
\pmb{\psi}=\pmb{\zeta}_{L}+\pmb{E}_{pos} 
\end{equation}
\begin{equation}
\bar{\pmb{d}}=Softmax(LN([\pmb{\psi}_{1};\pmb{\psi}_{2};...;\pmb{\psi}_{|\pmb{\upsilon}|}] cls^{\mathrm{T}})) 
\end{equation}

To train the model and learn an approximate distribution that can approximate the real distribution, we employ the symmetric Jensen-Shannon Divergence (JSD) distance as the loss function.
\begin{IEEEeqnarray}{c} 
Loss_{p\ or\ a}=JS(\bar{\pmb{d}}_{p\ or\ a}||\pmb{d}_{p\ or\ a})
\end{IEEEeqnarray}

Up until this point, we have discussed the structure of the NN. Refer to Algorithm 1 below for a detailed outline of the training process.
\begin{algorithm} 
  \caption{Distribution Learner Training}  
  \KwIn{
  Sampled data set of pairs $(\pmb{d},\pmb{\epsilon})$ \\

  }  
  \KwOut{Learnable parameter set of the converged Distribution Learner $F^{*} \to \Phi^{*}$
}  
\textbf{Initialize}  Divide the sampled data into training set $S_{t}$ and valid set $S_{v}$

  \For{$e=1;e \le epoch $}  
  {  
  
    \While{$S_{t} \neq \emptyset$}  
    {  
      Sample batch $S_{b}$ from $S_{t}$;\\
      Infer $\bar{\pmb{d}}$ for each pairs in $S_{b}$;\\
      Calculate average loss of $S_{b} \to {L_{b}}$ based on Eq(17) and update learnable parameter set $\Phi$ by BP algorithm;\\
      $S_{t}=S_{t} / \{S_{b}\}$;  
    } 
    Infer $\bar{\pmb{d}}$ for each pairs in $S_{v}$;\\
    Calculate average loss of  $S_{v} \to {L_{v}}$;\\
    Recover $S_{t}$;\\
  }  
  determine $e^{*}$ $\gets$ $argmin_{e \in \{1,...,epoch\}}L_{v}$;\\
  $\Phi^{*} \gets \Phi_{e^{*}}$ as the parameter set of converged Distribution Learner $F^{*}$;
\end{algorithm}

\subsection{Optimization}\label{AA}
The traditional method focuses on minimizing the difference between observed traffic counts and simulated traffic counts, which only aids numerical optimization between the estimated OD matrix and the real OD matrix and may get stuck in local optima. In order to improve the optimization process, we incorporate the best inference distribution $\pmb{d}^{*}$ into the optimization, and additionally, to retain features where the distribution is more pronounced, we utilize the reverse Kullback-Leibler Divergence (KLD) as the loss function.

\begin{IEEEeqnarray}{c} 
\min_{\hat{\pmb{t}}}R(\hat{\pmb{t}})=\min_{\hat{\pmb{t}}}(N(\hat{\pmb{t}})+S(\hat{\pmb{t}}))  \\
N(\hat{\pmb{t}})=\frac{\Theta}{2}(\pmb{\epsilon}-\hat{\pmb{\epsilon}})^{\mathrm{T}}(\pmb{\epsilon}-\hat{\pmb{\epsilon}}) \IEEEnonumber\\
S(\hat{\pmb{t}})=KL_{r}(\hat{\pmb{d}}_{p}||\pmb{d}^{*}_{p})+KL_{r}(\hat{\pmb{d}}_{a}||\pmb{d}^{*}_{a}) \IEEEnonumber\\
where \ 
\hat{\pmb{\epsilon}}=\pmb{P}\hat{\pmb{t}} \IEEEnonumber
 \end{IEEEeqnarray}

The balance factor $\Theta$ is used to weigh the relative importance of numerical and structural optimization during the optimization process. Since the numerical deviations are relatively large integers, while the structure is measured using Pearson similarity, the parameter $\Theta$ is used to adjust the scale between numerical optimization and structural optimization. For example, in our case, we set $\Theta$ to a range of values between 1e-6 and 1e-8. If $\Theta$ is set too small, the objective function will primarily optimize the numerical part, resulting in slow iterations. On the other hand, if $\Theta$ is set too large, the objective function will primarily optimize the structural part. Adjusting $\Theta$ to a magnitude close to the numerical part is sufficient.

Similar to Eq. 3, in order to enforce non-negativity constraints, we solve for the step size that minimizes $R$.

The optimization algorithm is shown in Algorithm 2.

\begin{algorithm}  
  \caption{Optimization Algorithm}  
  \KwIn{ 
 Observed traffic counts vector $\pmb{\epsilon}$\\
 Inferred OD distribution $\pmb{d}^{*}_{p}$ and $\pmb{d}^{*}_{a}$ from converged Distribution Learner $F^{*}_{p}$ and $F^{*}_{a}$ based on Eq(6), respectively.
  }  
  \KwOut{Best estimated OD matrix $\pmb{t}^{*}$
}
\textbf{Initialize}  Balance factor $\Theta$, Scaling factor $\alpha$, Initialized OD matrix $\tilde{\pmb{t}}$, set $\hat{\pmb{t}}^{k}=\tilde{\pmb{t}}$ and $k=0^{th}$ iteration; \\
\Repeat{$R(\hat{\pmb{t}}^{k})$ convergent}  
{  
 \textbf{Lower level:} Parallel simulate $\hat{\pmb{t}}^{k}$  with the simulator and obtain the simulated traffic counts vector $\hat{\pmb{\epsilon}}$. Assignment matrix $\pmb{P}$ is an average estimated using a back-calculation procedure;\\ 
 
 \textbf{Upper level:} $\hat{\pmb{t}}^{k+1}=\min_{\hat{\pmb{t}}^{k}}R(\hat{\pmb{t}}^{k})$ based on Eq(2);\\
  }
\end{algorithm} 


\section{EXPERIMENTS AND RESULTS}
We conducted experiments on two datasets. Firstly, we experimented on synthetic traffic data in a large-scale real city and tested different OD divisions for $15\times 15$ and $50\times 50$ to evaluate the generalization performance of our NN model and optimization estimator. Secondly, we conducted experiments on real traffic data in a ring-road region and tested different initial OD matrices and assignment algorithms (Dijkstra and UE) to verify the stability performance on actual traffic data. Furthermore, to demonstrate the efficacy of our NN model design, we conducted parametric analysis and ablation experiments. We implemented our experiments using Python programming language and the TensorFlow system \cite{abadi2016tensorflow} for NN modeling and gradient calculation. We used the Sklearn library \cite{pedregosa2011scikit} for clustering algorithm and Scipy \cite{virtanen2020scipy} for constrained optimization.

\subsection{OD estimating evaluation and reference models}\label{AA}

Similar to \cite{behara2020novel}, we employed two indicators to evaluate the accuracy of estimated OD matrices against the real OD matrices.
\begin{itemize}
\item $RMSN(\hat{\pmb{t}},\pmb{t})$ \cite{antoniou2004incorporating}:
The $RMSN$ indicator measures the root mean square deviation between the estimated and real OD matrices. It is computed as the square root of the sum of squared differences between corresponding elements in the two matrices, normalized by the sum of elements in the real OD matrix. Symbolically, we can express it as:

\begin{equation}
RMSN(\hat{\pmb{t}},\pmb{t})=\frac{\sqrt{|\pmb{\upsilon}|^2\sum\limits^{|\pmb{\upsilon}|^2}_{i=1}(\pmb{t}_{i}-\hat{\pmb{t}}_{i})^2}}{\sum\limits^{|\pmb{\upsilon}|^2}_{i=1} \pmb{t}_{i}}
\end{equation}

\item $\rho(\hat{\pmb{t}},\pmb{t})$ \cite{djukic2013reliability}:
The $\rho$ indicator, on the other hand, measures the similarity in the structures of the estimated and real OD matrices. It is computed as the inner product of the mean-centered estimated and real OD matrices, divided by the product of their Euclidean norms. Symbolically, we can express it as:

\begin{equation}
\rho(\hat{\pmb{t}},\pmb{t})=\frac{(\pmb{t}-\pmb{\mu})^\mathrm{T}(\hat{\pmb{t}}-\pmb{\hat{\mu}})}{\sqrt{(\pmb{t}-\pmb{\mu})^{\mathrm{T}}(\pmb{t}-\pmb{\mu})} \sqrt{(\hat{\pmb{t}}-\pmb{\hat{\mu}})^{\mathrm{T}}(\hat{\pmb{t}}-\pmb{\hat{\mu}})} }  
\end{equation}

Here, $\pmb{\mu} \in \mathbb{R}^{|\pmb{\upsilon}|^{2}}_{\geq{0}}$ is a vector with each element value equal to the mean of $\pmb{t}$, and $\pmb{\hat{\mu}}$ correspond to the mean-centered $\hat{\pmb{t}}$.
\end{itemize}

We implement one traditional model for the performance comparison.

\begin{itemize}
\item $Traditional$: It solely focus on optimizing the numerical aspect, as exempliﬁed by the technique presented in part (B) of section 3. The optimization of the OD matrix is achieved through the application of the least square method, with the goal of minimizing traffic counts deviation.
\end{itemize}

\subsection{Parameter settings}\label{AA}

{\centering

\begin{tabular}{ll}
    \hline
	head number $h$ & 6 \\
	encoder layer $N$ & 4 \\
	learning rate $r$ & 5E-4 \\
	dimention $d$ & 128\\
	maximum trip value $M$ & 100\\
	density $\varepsilon$ & 0.015 \\
	degree of non-uniformity $\beta$ & 0.5\\
	$\Theta$  & 1e-6 to 1e-8 \\
    \hline

\end{tabular}

}

\subsection{Cologne}\label{AA}
\subsubsection{Study Network}\label{AA}
For our experiments on a large-scale city network, we selected the 400 $km^2$ area around Cologne, Germany, as our study region \cite{uppoor2013generation}. To simulate traffic flow, we utilized the simulator software SUMO. The network in our study region comprised of 31,584 intersections and 71,368 road sections, as shown in Fig. 3(a). To cluster the origin-destination (OD) nodes based on their Euclidean distance on the network, we employed the K-means algorithm \cite{arthur2006k}. For instance, when we selected the number of OD nodes $|\pmb{\upsilon}|=50$, which are shown in Fig. 3(b). Furthermore, we aggregated the road sections from 5,030 to 237, and each road section was directed.

\subsubsection{Dataset}\label{AA}
Two parameters are used to generate the OD matrix, namely the maximum trip value $M$ and the density $\varepsilon$. We employ a uniform random distribution ranging from 0 to $M=100$, and $\varepsilon=0.015$, resulting in a total of approximately 1875 trips for each sampled OD matrix. Our dataset consists of around 32K sampled data pairs, which are divided into a training set and a validation set in an 8:2 ratio. The training set is utilized to train a model for inference capability, whereas the validation set is employed to assess the generalization performance of the model. The test set serves as the ground truth and includes the matrices listed in Table 1.

Refer to \cite{uppoor2013generation}, the ground truth data is synthetically generated to closely resemble the actual traffic conditions in urban areas. To ensure that the learned distribution is stable across different numerical quantities, we test the model on three time periods, namely 6:00-6:10, 6:00-6:30, and 6:00-7:00 o'clock, which have similar distributions but different traffic flow. Furthermore, we select six different time periods, namely 6:00-7:00, 9:00-10:00, 12:00-13:00, 15:00-16:00, 18:00-19:00, and 21:00-22:00 o'clock, to verify the stable effect on different distributions.

\begin{table}
\centering
	\caption{OD matrices features}
	\label{table}
	\setlength{\tabcolsep}{3pt} 
	\renewcommand\arraystretch{1.5} 
	\begin{tabular}{m{1.25cm}<{\centering}|m{1.25cm}<{\centering}|m{1cm}<{\centering}m{1cm}<{\centering}|m{1cm}<{\centering}m{1cm}<{\centering}|m{1cm}<{\centering}}
	\toprule
		 \multirow{2}{*}{OD matrix}&\multirow{2}{*}{Total travels}   & \multicolumn{2}{c|}{ $KL_{r}(\pmb{d}_{p}||St)$} & \multicolumn{2}{c|}{$KL_{r}(\pmb{d}_{a}||St)$} & \multirow{2}{*}{Density}  \\
		(o'clock)& & 15 $\times$ 15 & 50 $\times$ 50 & 15 $\times$ 15 & 50 $\times$ 50    \\ \hline 
		$6-6:10$ & 10876 &0.4854& 0.8004 &1.0207& 1.2695 & 0.2468  \\ \hline
		$6-6:30$  & 38375 &0.4867& 0.7035 &1.0559& 1.3046 & 0.3152  \\ \hline
		$6-7$       & 97124 &0.4909& 0.7994 &1.0553& 1.3119 & 0.3568  \\ \hline
		$9-10$     & 44398 & 0.5569&0.8699 &0.7803& 1.0593 & 0.3052  \\ \hline
		$12-13$   & 67702 &0.7883 &1.0969 & 0.6047&0.9303 & 0.3408  \\ \hline
		$15-16$   & 99228 &0.8018 &1.0820 & 0.5682&0.9002 & 0.3600  \\ \hline
		$18-19$   & 78222 &0.7497 &1.0376 &0.5542 &0.8751 & 0.3448  \\ \hline
		$21-22$   & 27144 &0.7597 &1.0420 &0.5186 &0.8486 & 0.2908  \\ 
		\bottomrule

	\end{tabular}
	\label{tab1}
\end{table}

\begin{figure}
\centering
\subfigure[Cologne city network]{\includegraphics[width=4.3cm]{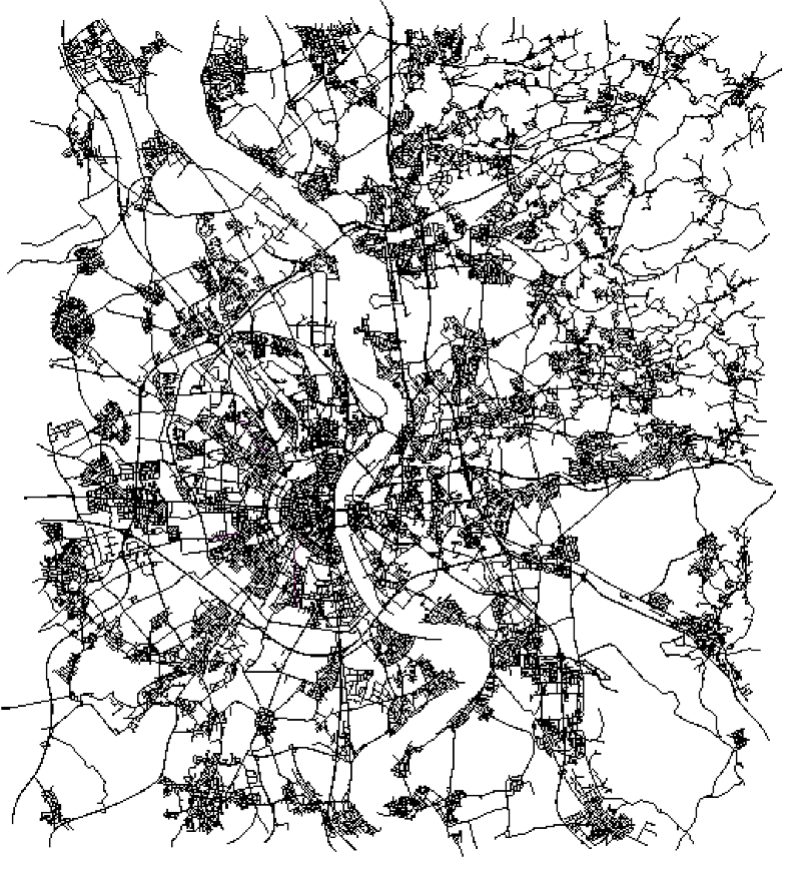}}
\subfigure[nodes and road sections]{\includegraphics[width=4.3cm]{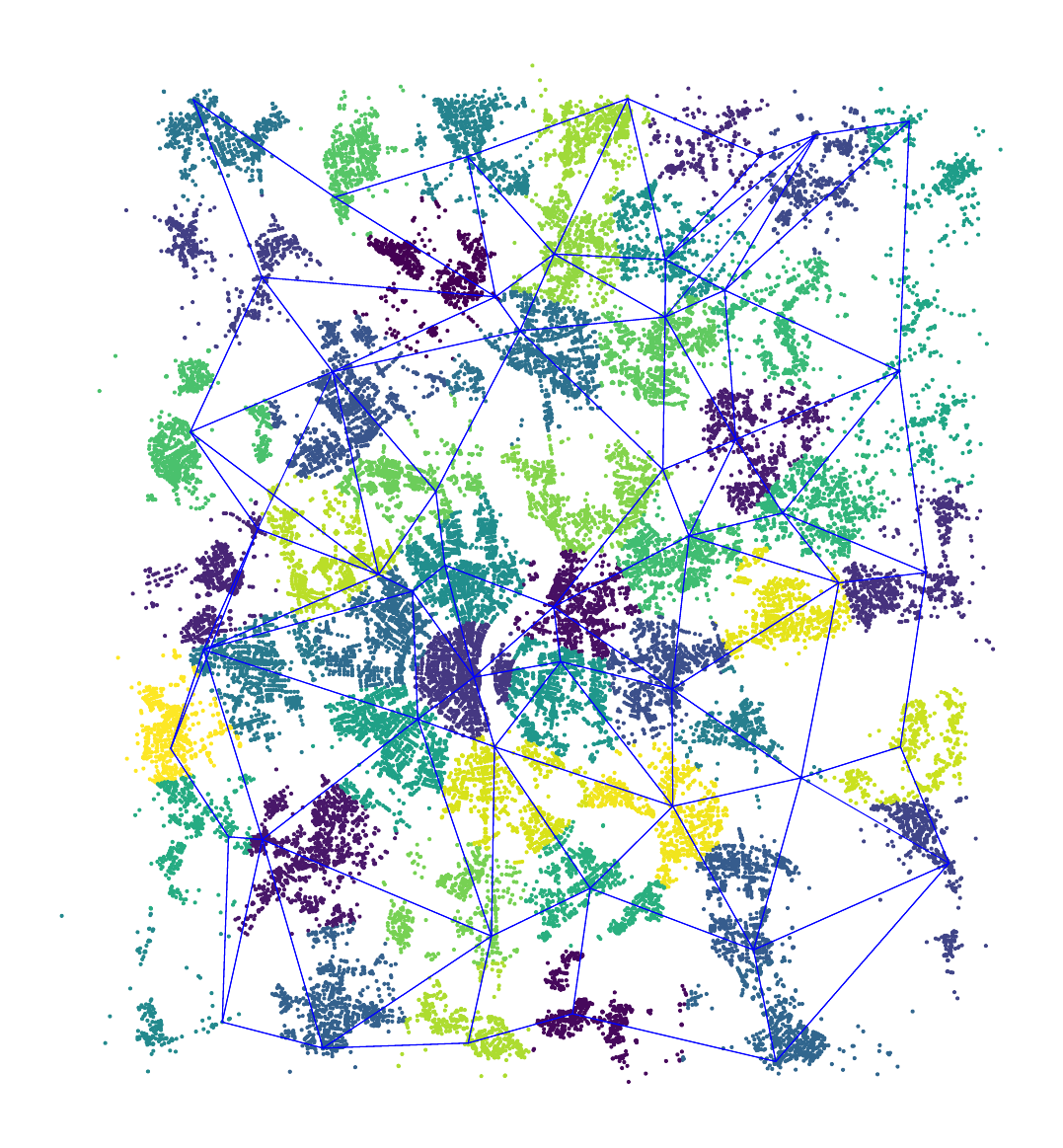}}
\caption{The aggregate traffic counts operation is denoising because vehicles will choose different road sections between two connected adjacent OD nodes.}
\label{fig:2 }  
\end{figure}

\begin{figure}
\centering
\subfigure[production distribution on $15 \times 15$]{\includegraphics[width=4.3cm]{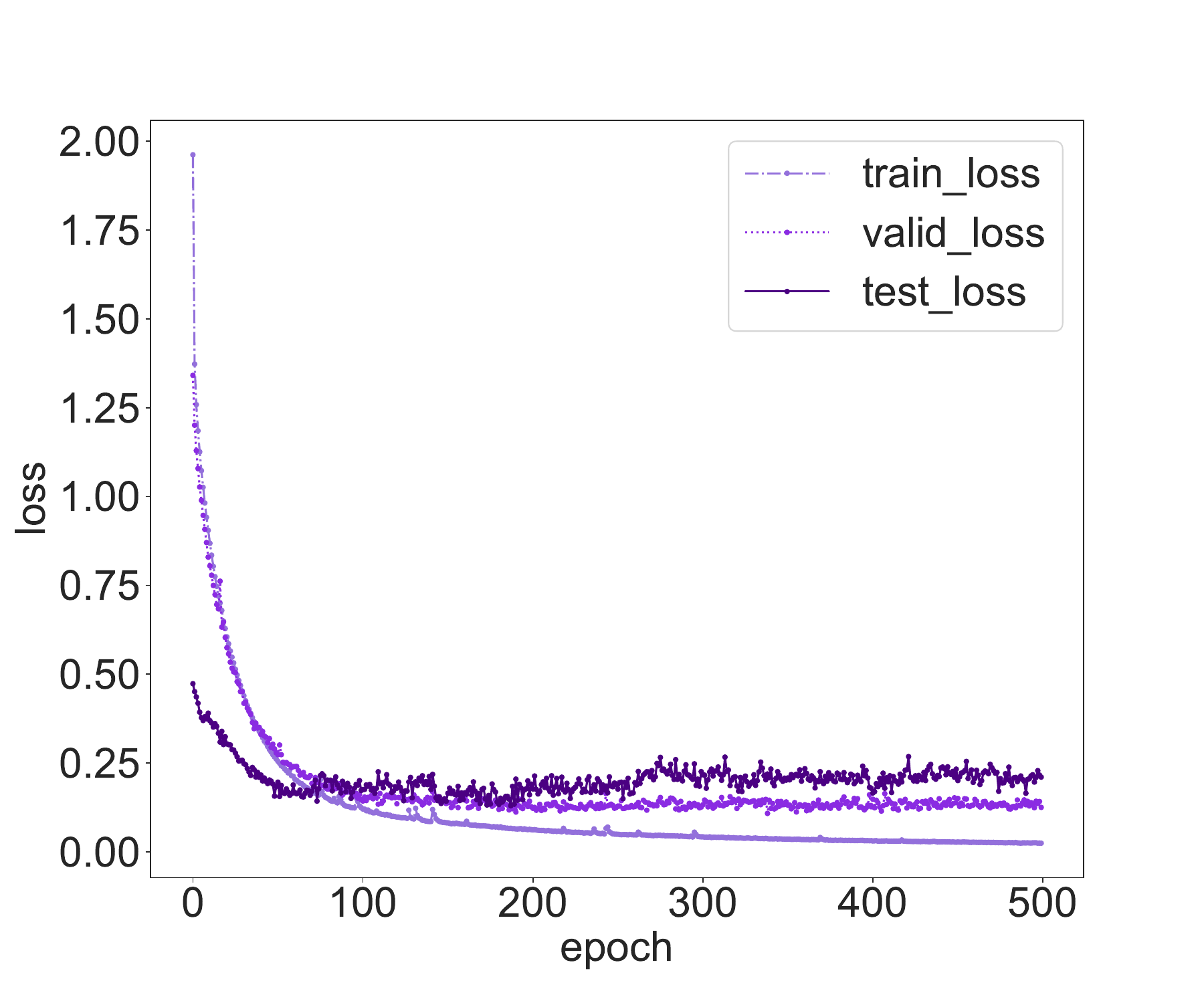}}
\subfigure[production distribution on $50 \times 50$]{\includegraphics[width=4.3cm]{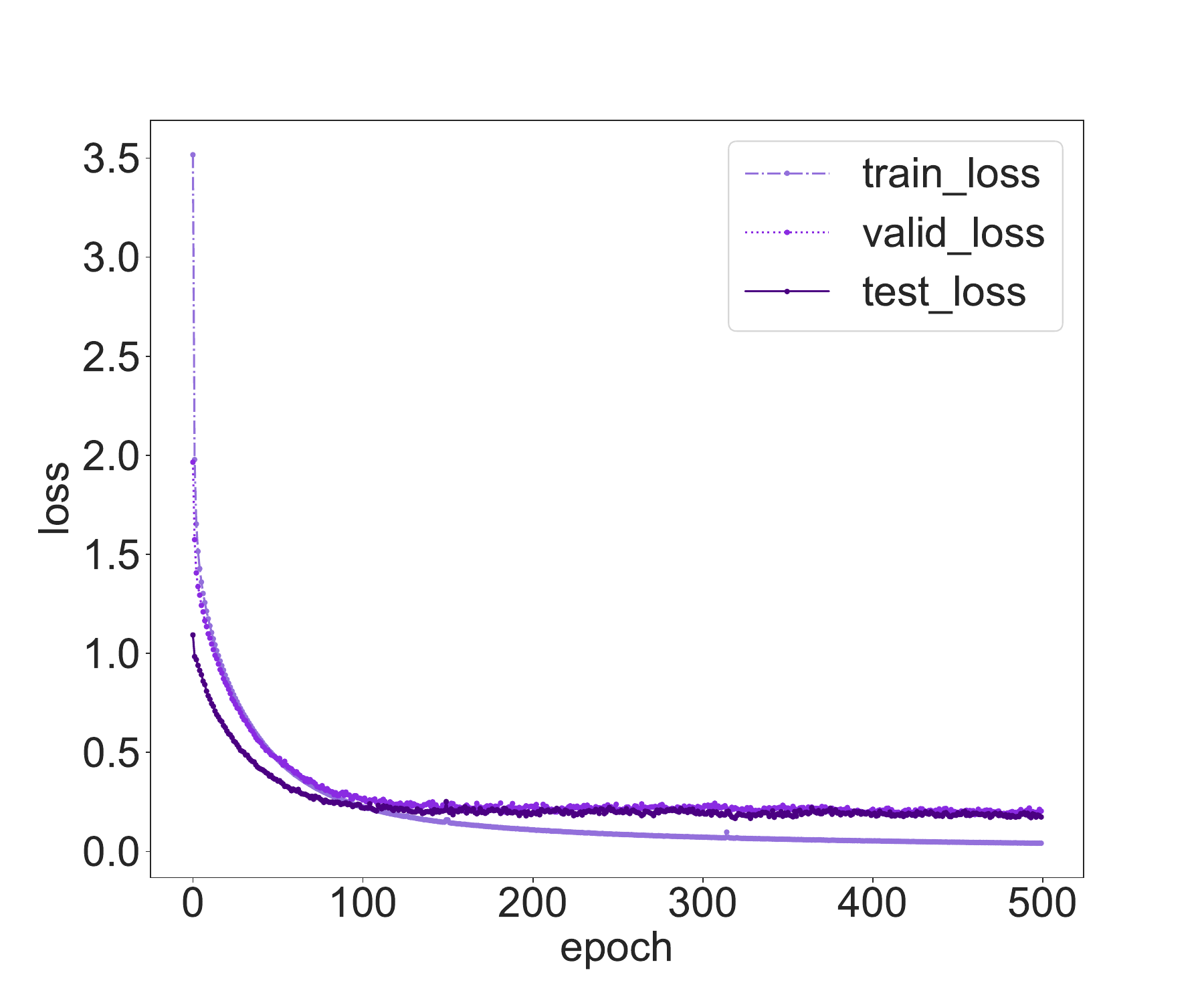}}
\caption{Learning curves, the reverse KLD learning curve $KL_{r}(\bar{\pmb{d}}||{\pmb{d}})$ can indicate how well deep learning fits the features of the high-probability part of the real distribution} 
\label{ }  
\end{figure}

\subsubsection{Results and analyze}\label{AA}

During the training phase, we observe that the reverse KLD learning curve of both the test set and the validation set are consistent (as depicted in Fig. 4). Based on this, we conclude that for the distribution sampling, the test set is surrounded by the sampled samples, which effectively avoids the issue of numerical sampling. Numerical sampling may result in failure to surround the ground truth, causing the test set to become an outlier in training and fail to learn.

Moving onto the inference phase, as shown in Table 2, the distribution learner takes the traffic count distribution as input and generates the corresponding matrix's production and attraction distribution. A lower reverse KLD indicates that the inferred distribution better captures the higher probability parts of the real distribution. The comparison between the approximate distribution in blue and the real distribution in gray, visualized in Fig. 5 for two different periods 6-7 o'clock and 18-19 o'clock, indicates that the parts with high probability are predicted correctly. Additionally, the production distribution predicted by each node based on the sections exiting from nodes at 6-7 o'clock is similar to the attraction distribution predicted by each node based on the sections entering nodes at 18-19 o'clock due to the symmetrical distribution of morning and evening peaks.

During the optimization phase, it is essential to begin with an unbiased initial matrix that does not incorporate any prior knowledge. To achieve this, we set all elements of the matrix to 30, with the exception of the diagonal, which is set to 0. In our initial case for a $15 \times 15$, we found that incorporating the inferred distribution as a guide led to significant improvements in the estimation of OD matrices compared to traditional methods. Fig. 6 displays the iterative process of each indicator of OD matrix 18-19 o'clock on $15 \times 15$. As shown in Fig. 6(a) and 6(b), our method accelerates the optimization process.  From Fig. 6(c), it is noteworthy that although our approach did not always achieve a better distribution than the traditional method, our guided optimization process still produced superior convergence results. Specifically, our method's $KL_{r}(\hat{\pmb{d}}_{p}||\pmb{d}_{p})$ and $KL_{r}(\hat{\pmb{d}}_{a}||\pmb{d}_{a}) $ values were 0.1082 and 0.0787, respectively, while the traditional method's values were 0.0617 and 0.0618. However, our method's $RMSN$ and $\rho$ values of 1.8957 and 0.7370, respectively, still outperformed those of the traditional method, which were 2.0394 and 0.6848. On the contrary, in the case of 9-10 o'clock, our method introduced excessive constraint deviation from inferred $KL_{r}(\pmb{d}_{p}^{*}||\pmb{d}_{p})$ and $KL_{r}(\pmb{d}_{a}^{*}||\pmb{d}_{a})$ values of 0.1536 and 0.1515, resulting in a final outcome that was inferior to the traditional approach. Fig. 6(d) presents the scatter plot of the actual OD matrix (x-axis) and the optimized OD matrix (y-axis), with a closer scatter distribution to the x-y line indicating a closer optimization result to the actual result. To provide a clearer explanation, we calculated the $R^{2}$ metrics for three results $traditional$, $ours$, and $best$, in comparison to the ground truth values. The $R^{2}$ values obtained are 0.441, 0.517, and 0.631, respectively. A higher $R^{2}$ value indicates that the estimated values' distribution is closer to the ground truth values.

Overall, our results suggest that incorporating distribution information as a guide can improve the estimation of OD matrices, even when the inferred distribution is not more accurate than the traditional method.

\begin{figure*}
\centering
\subfigure[$\pmb{d}_{p}$ of 6-7 o'clock]{\includegraphics[width=4.4cm]{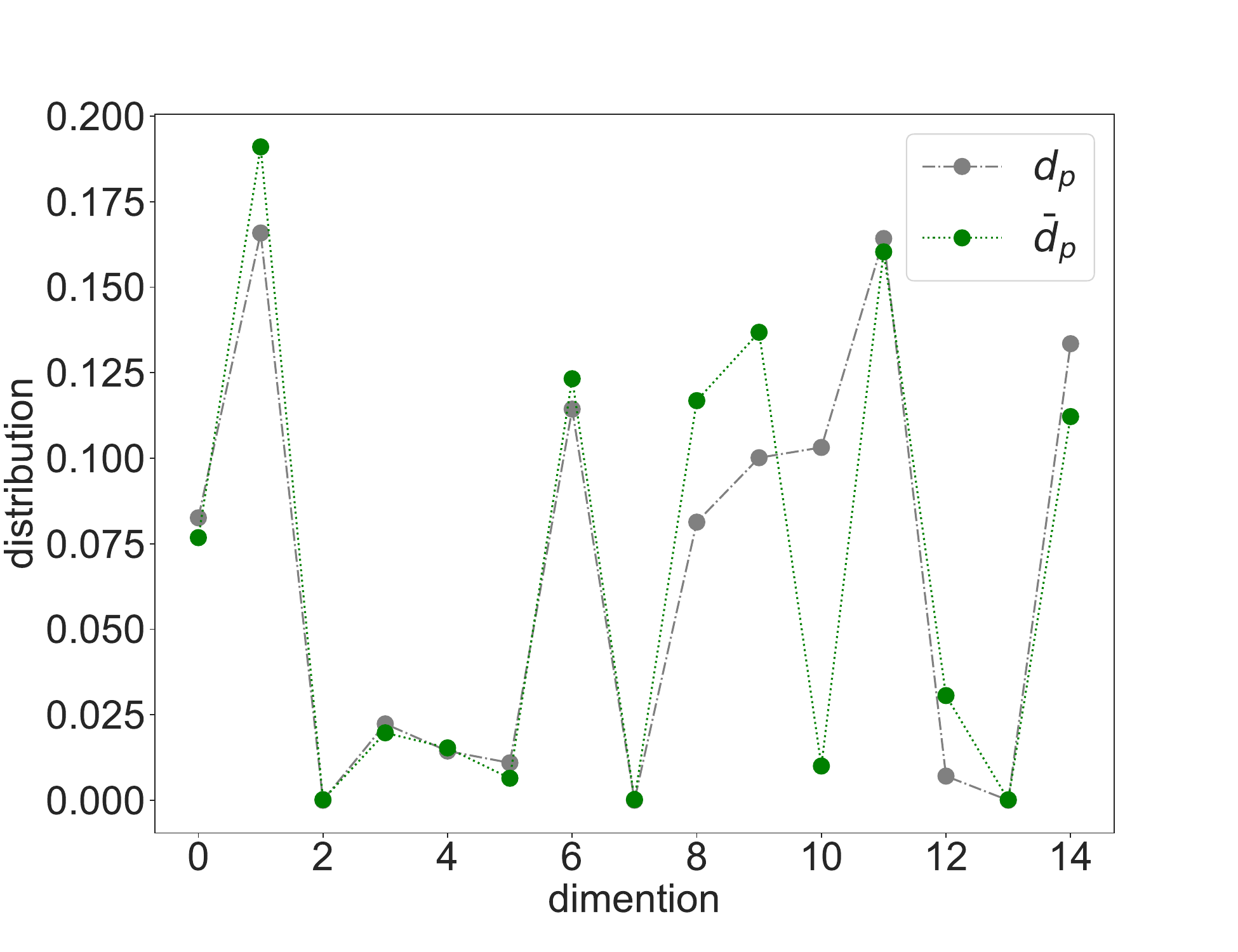}}
\subfigure[$\pmb{d}_{a}$ of 6-7 o'clock]{\includegraphics[width=4.4cm]{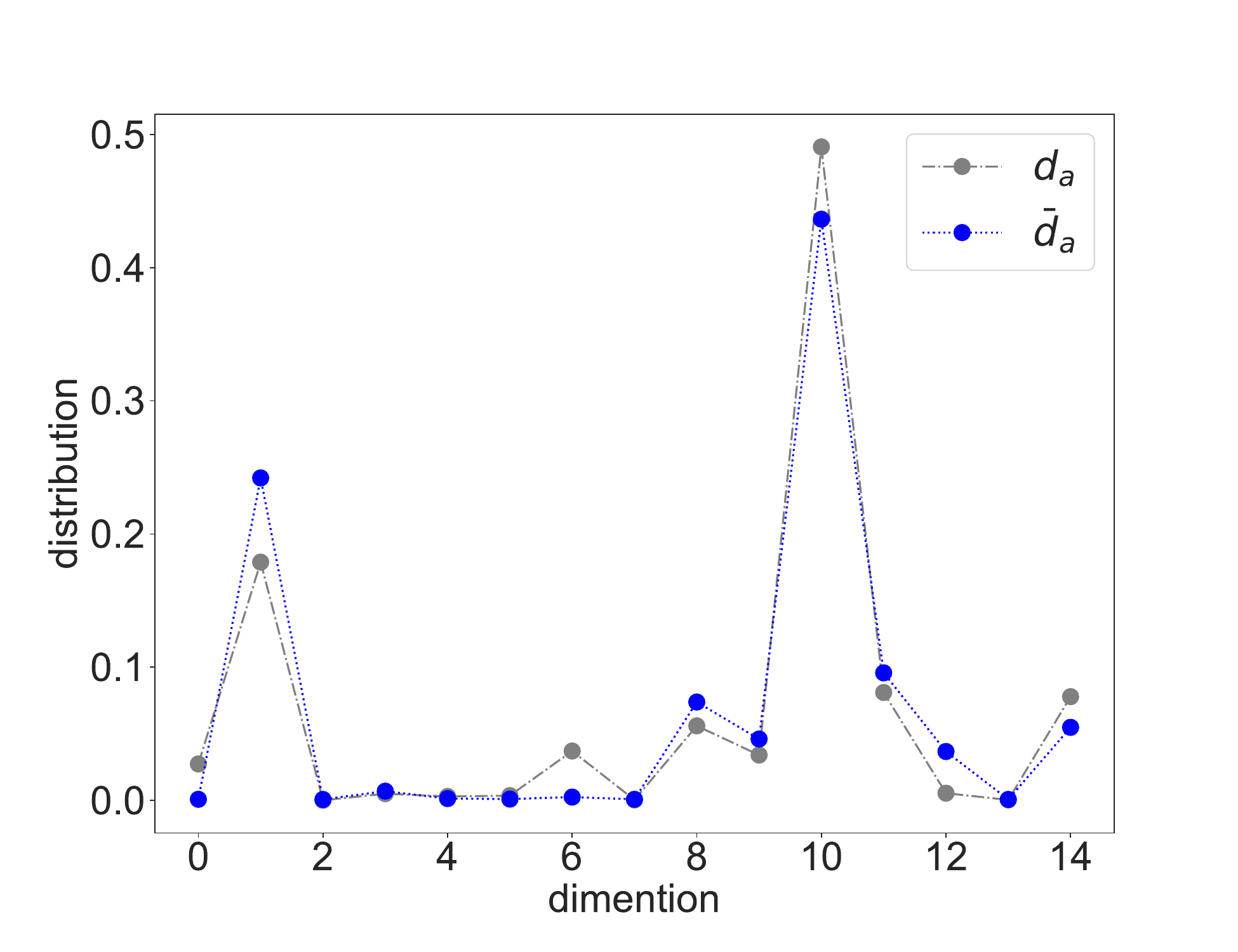}}
\subfigure[$\pmb{d}_{p}$ of 18-19 o'clock]{\includegraphics[width=4.4cm]{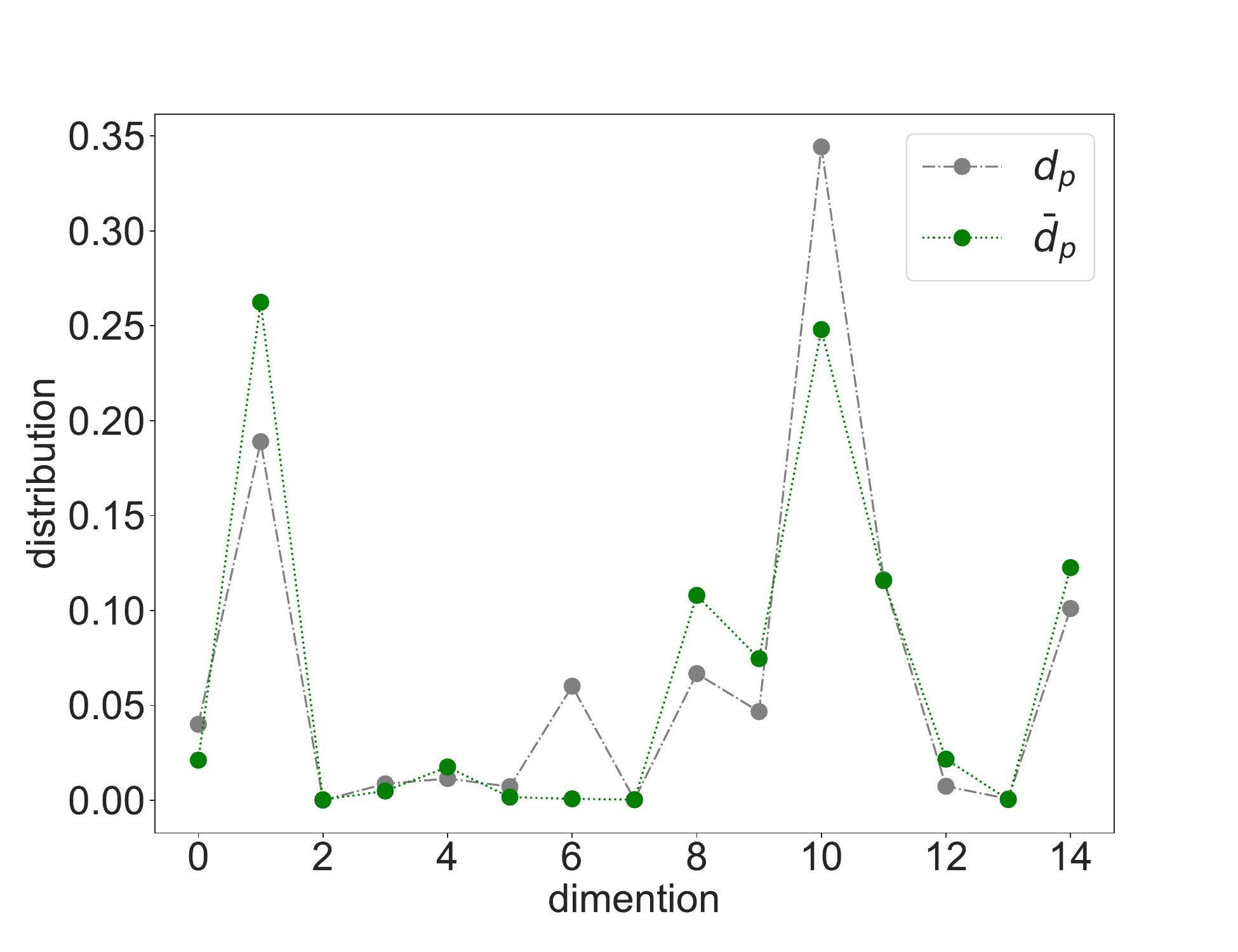}}
\subfigure[$\pmb{d}_{a}$ of 18-19 o'clock]{\includegraphics[width=4.4cm]{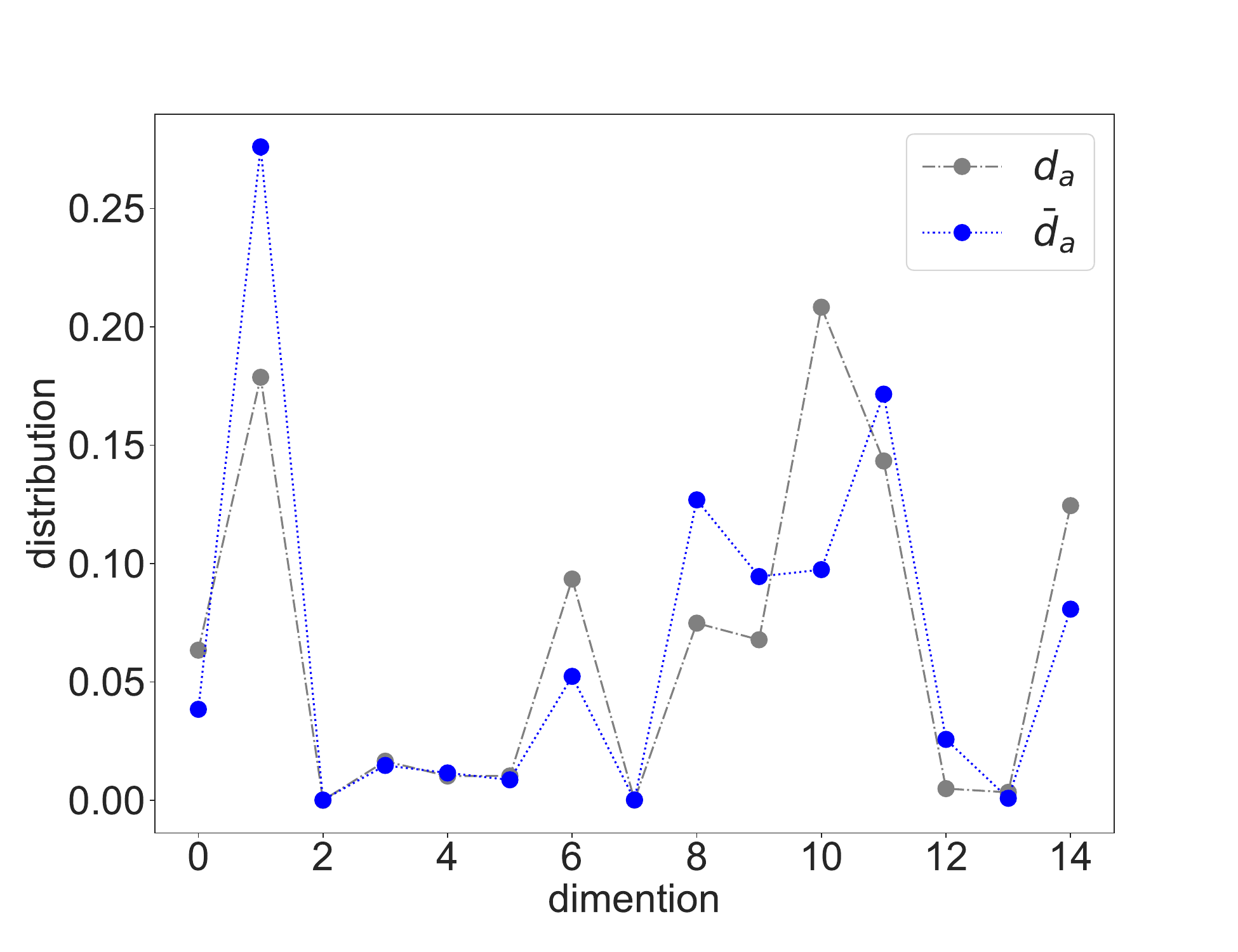}}

\caption{Comparison of inferred and true distributions. Example: 6-7 o'clock and 18-19 o'clock on $15 \times 15$} 
\label{}  
\end{figure*}

\begin{table*}
\centering
	\caption{Inference result}
	\label{table}
	\setlength{\tabcolsep}{3pt} 
	\renewcommand\arraystretch{1.2} 
	\begin{tabular}{m{2cm}<{\centering}|m{1.5cm}<{\centering}|m{1.5cm}<{\centering}|m{1.5cm}<{\centering}|m{1.5cm}<{\centering}|m{1.5cm}<{\centering}|m{1.5cm}<{\centering}|m{1.5cm}<{\centering}|m{1.5cm}<{\centering}|m{1.5cm}<{\centering}}
	\toprule
		 
		 \multicolumn{2}{c|}{OD matrix o'clock}&$6-6:10$&$6-6:30$&$6-7$&$9-10$&$12-13$&$15-16$&$18-19$&$21-22$\\ \hline
		
		\multirow{2}{*}{$KL_{r}(\pmb{d}_{p}^{*}||\pmb{d}_{p})$}& 15 $\times$ 15 &0.1183  & 0.1185 & 0.1113 &   0.1536&   0.1487& 0.1502 & 0.1256 &   0.1215  \\
		&50 $\times$ 50 &0.1686  & 0.1562 &0.1615  & 0.2078 & 0.1221 &0.1436  &0.1159  & 0.0838  \\ \hline
		
		\multirow{2}{*}{$KL_{r}(\pmb{d}_{a}^{*}||\pmb{d}_{a})$}& 15 $\times$ 15 & 0.1207 &   0.1190& 0.1213 & 0.1515 & 0.1236 & 0.1503 & 0.1311  &0.0671   \\
		&50 $\times$ 50 &0.1125  &0.1067  &0.0981  &0.1372  &0.1753  &0.1834  &0.1668  & 0.1511  \\ 
		\bottomrule

	\end{tabular}
	\label{tab1}
\end{table*}

\begin{figure*}
\centering
\subfigure[$RMSN$]{\includegraphics[width=4.4cm]{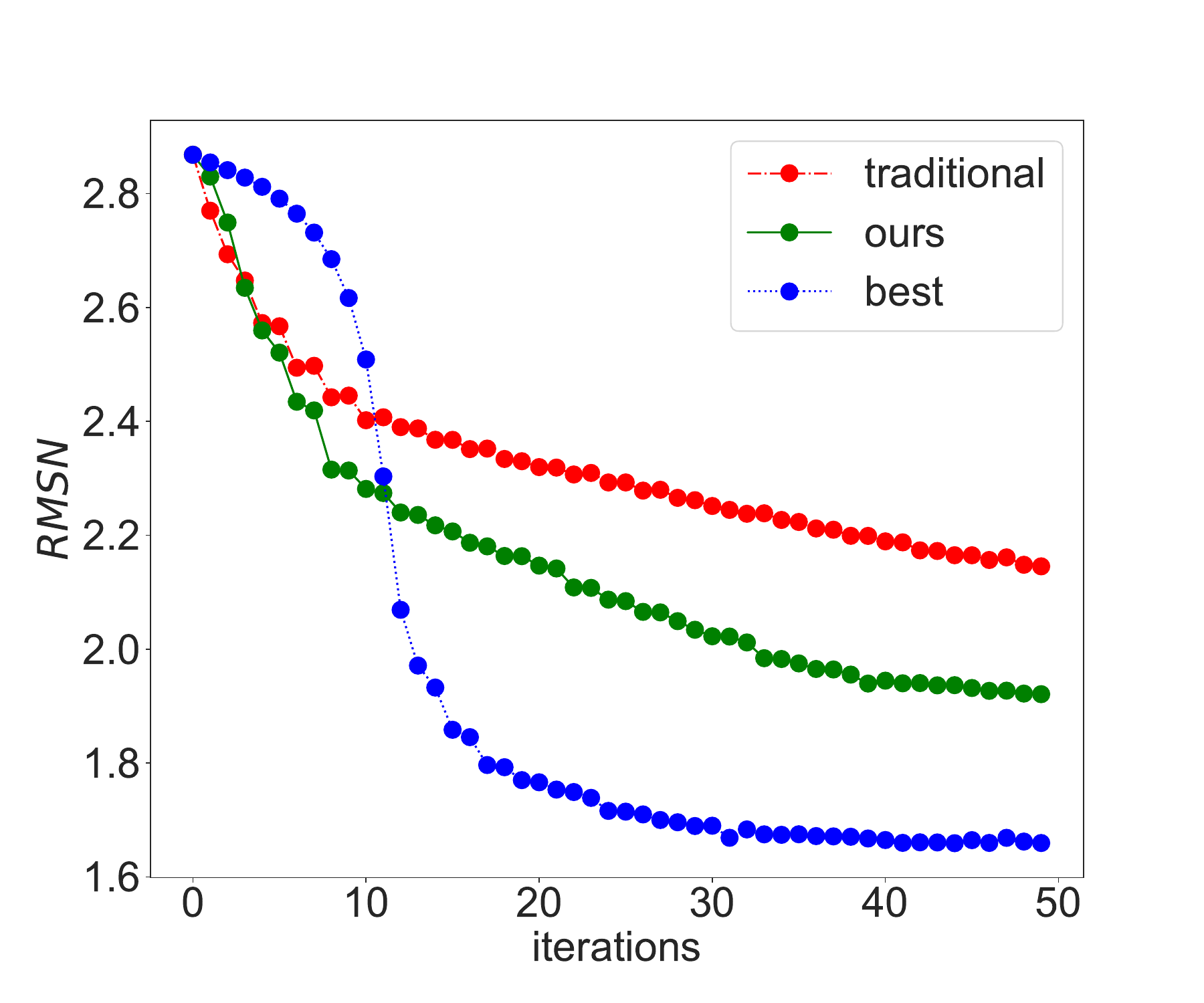}}
\subfigure[$\rho$]{\includegraphics[width=4.4cm]{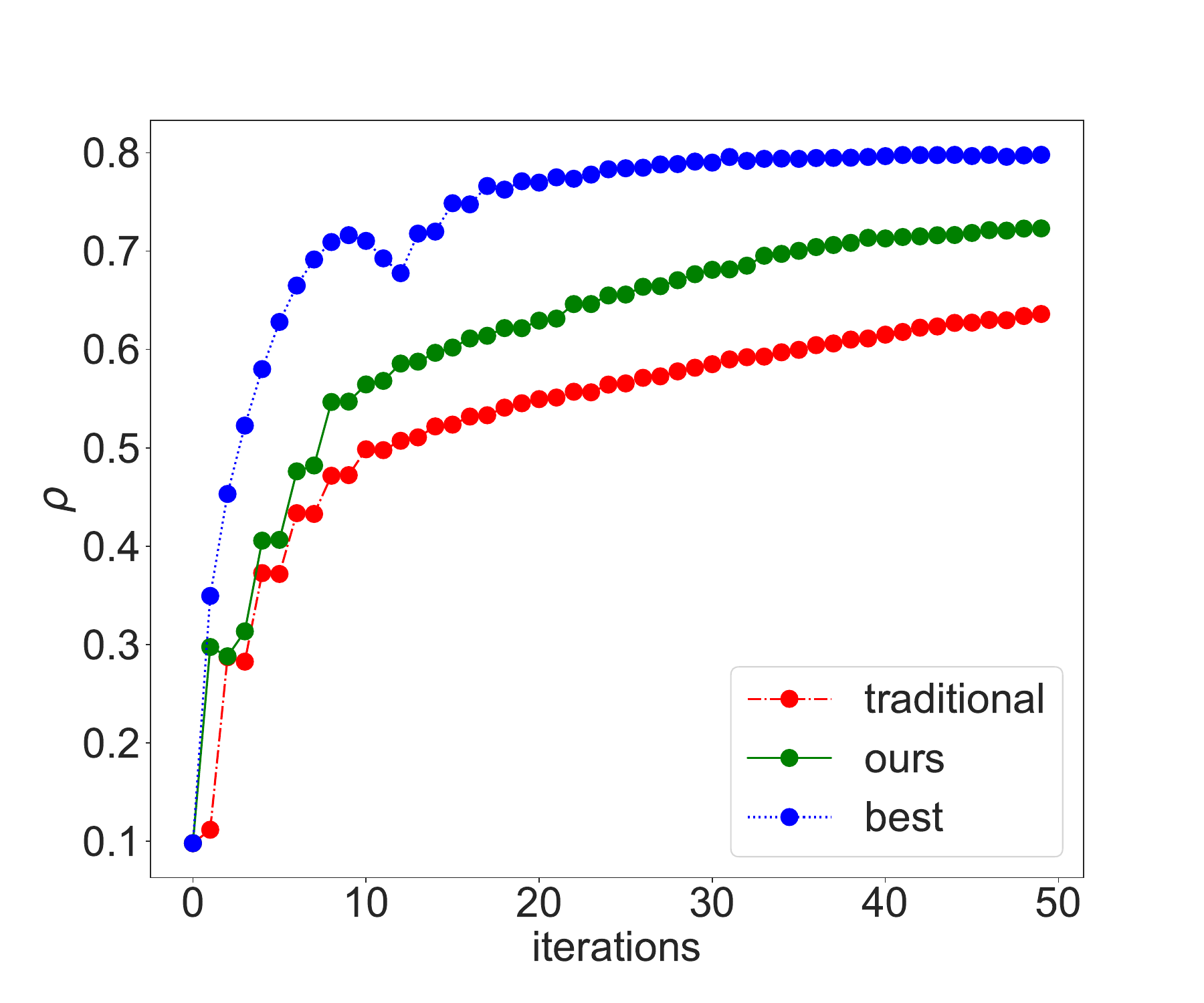}}
\subfigure[$KL_{r}(\hat{\pmb{d}}||\pmb{d})$]{\includegraphics[width=4.4cm]{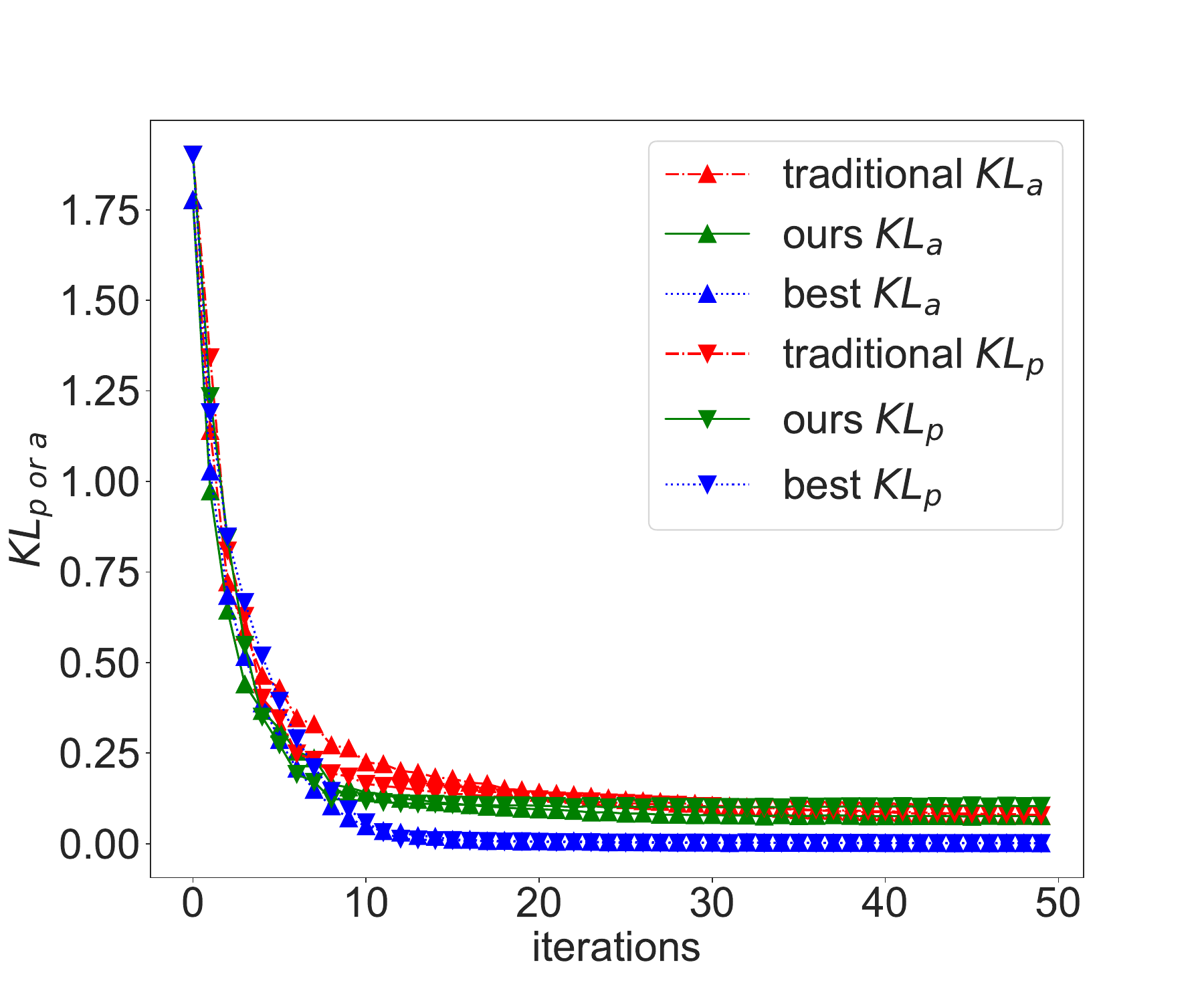}}
\subfigure[x-y]{\includegraphics[width=4.4cm]{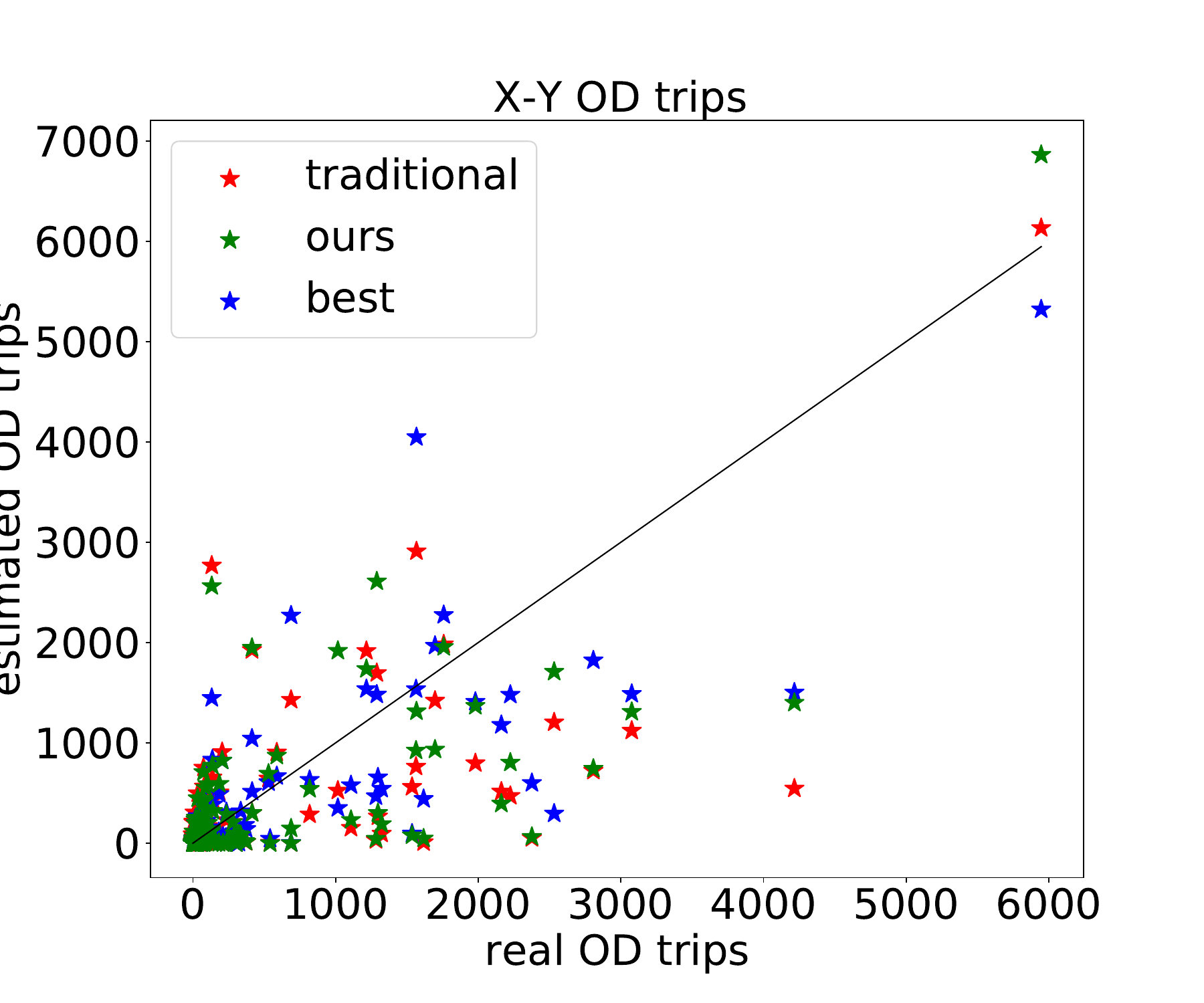}}

\caption{Iterative process and indicators results. Example: 18-19 o'clock on $15 
\times 15$} 
\label{ }  
\end{figure*}

\begin{table*}
\centering
	\caption{Optimization result}
	\label{table}
	\setlength{\tabcolsep}{3pt} 
	\renewcommand\arraystretch{1.2} 
	\begin{tabular}{m{1cm}<{\centering}|m{2cm}<{\centering}|m{1.5cm}<{\centering}|m{1.5cm}<{\centering}|m{1.5cm}<{\centering}|m{1.5cm}<{\centering}|m{1.5cm}<{\centering}|m{1.5cm}<{\centering}|m{1.5cm}<{\centering}|m{1.5cm}<{\centering}}
	\toprule
		 
		 \multicolumn{2}{c|}{OD matrix(o'clock)}&$6-6:10$&$6-6:30$&$6-7$&$9-10$&$12-13$&$15-16$&$18-19$&$21-22$ \\ \hline

		&\multicolumn{8}{c}{$KL_{r}(\hat{\pmb{d}}_{p}||\pmb{d}_{p})$} \\   
		\multirow{2}{*}{15 $\times$ 15}&Traditional & 0.0965 &0.0939  &0.0870  & 0.0703 & 0.0542 & 0.0595 & 0.0617 &0.0829    \\ 
		&Ours & 0.1114  & 0.1184 &  0.1051 &0.1434  & 0.1103  & 0.1390 & 0.1082 & 0.1177  \\ 
		
		\multirow{2}{*}{50 $\times$ 50}&Traditional & 0.9646 & 0.6939 & 0.6893 & 0.3181 & 0.2489 & 0.2091 & 0.2260 & 0.3928   \\ 
		&Ours & 0.5843 & 0.1529 & 0.1561 & 0.2072 & 0.1284 & 0.1940 & 0.1381 & 0.0850 \\ \hline
		
		&\multicolumn{8}{c}{$KL_{r}(\hat{\pmb{d}}_{a}||\pmb{d}_{a})$} \\   
		\multirow{2}{*}{15 $\times$ 15}&Traditional &0.0760  & 0.0565 & 0.0622 &0.0623  & 0.0581  &  0.0621 & 0.0618 &0.0823   \\ 
		&Ours & 0.1115 & 0.1064 &0.1017  &0.1534  & 0.0697 &0.1207  &0.0787  &0.0687   \\ 

		\multirow{2}{*}{50 $\times$ 50}&Traditional & 1.1167 & 0.7346 & 0.5572 & 0.3077 & 0.2643 & 0.2454 & 0.2511 & 0.3996   \\ 
		&Ours & 0.7455 & 0.1095 & 0.1031 & 0.1363 & 0.1536 & 0.1899 & 0.1500 & 0.1471  \\ \hline

		 &\multicolumn{8}{c}{$RMSN$} \\   
		\multirow{2}{*}{15 $\times$ 15}&Traditional & 1.5218 &1.8241  &  1.9663&  1.8340 &   2.0157&2.1327  & 2.0394 &1.7953   \\ 
		&Ours &  \bfseries 1.4561& \bfseries 1.7399& \bfseries1.8753 &1.9925  &\bfseries  1.8659 & \bfseries1.9547 &\bfseries1.8957  &\bfseries 1.7783 \\ 

		\multirow{2}{*}{50 $\times$ 50}&Traditional & 2.4670 & 3.0368 & 3.7091 & 3.5269 & 3.7847 & 3.826 & 3.5257 & 3.0357  \\ 
		&Ours & \bfseries2.4106 & \bfseries2.9748 & \bfseries3.4233 & 3.5420 & \bfseries3.7251 & \bfseries3.7420 & \bfseries3.4455 & \bfseries2.7715 \\ \hline
		 
		&\multicolumn{8}{c}{$\rho$} \\   
		\multirow{2}{*}{15 $\times$ 15}&Traditional & 0.8704 & 0.8089 & 0.7781 & 0.7652 &  0.7401&  0.6865&0.6848  & 0.7477   \\ 
		&Ours &  \bfseries0.8705  & \bfseries0.8271 & \bfseries0.7925 &  0.7155&\bfseries 0.7783 &\bfseries0.7305  &\bfseries 0.7370 &\bfseries0.7579  \\ 

		\multirow{2}{*}{50 $\times$ 50}&Traditional & 0.8299 & 0.7405 & 0.6053 & 0.6324 & 0.5887 & 0.5519 & 0.5917 & 0.6743   \\ 
		&Ours & \bfseries0.8391 & \bfseries0.7583 & \bfseries0.6848 & \bfseries0.6370 & \bfseries0.6289 & \bfseries0.5878 & \bfseries0.6337 & \bfseries0.7370 \\
		\bottomrule

	\end{tabular}
	\label{tab1}
\end{table*}

When dealing with a larger number of OD partitions, the optimizer's performance deteriorates due to the exponential increase in variables. To evaluate the effectiveness of our method on a larger scale, we conducted tests on a $50 \times 50$ case. Upon analyzing the combined data from Table 2 and Table 3, it became evident that our inferred distribution showed greater similarity to the actual distribution compared to traditional methods such as OD matrix 6-7 o'clock. The traditional methods resulted in $KL_{r}(\hat{\pmb{d}}_{p}||\pmb{d}_{p})$ and $KL_{r}(\hat{\pmb{d}}_{a}||\pmb{d}_{a})$ values of 0.6893 and 0.5572, respectively, while our inferred distribution yielded values of 0.1615 and 0.0981 for the same measures.

The results presented in Table 3 indicate that our approach enhances all OD matrices during optimization phases on $50 \times 50$, as the inferred distribution more closely resembles the actual distribution compared to the distribution obtained through traditional optimization methods. However, we discovered that in the case of the $50 \times 50$, continuing to optimize the numerical part after the structural part converges makes the $RMSN$ and $\rho$ worse. Since in the case, the distribution provides fewer constraints, the accuracy of the inferred distribution provided by us is not currently high. So, a more accurate distribution is required to ensure consistency between the numerical optimization and structural optimization. In our experiments, we removed the numerical optimization component (by setting $\Theta=0$) after the structural optimization converged.

We incorporated the actual distribution into our optimization method is marked as $best$ with the blue line in Fig. 6(a), (b) and (c), to illustrate the upper limit of our method. And we took the average of the indicators of the 8 matrices and observed the results. For $15 \times 15$ case, the average $RMSN$ and average $\rho$ of the $best$ method were 1.5586 and 0.8438. When compared to the $traditional$ 1.8911 and 0.7602, it showed improvement in $\mathbf{17.58 \%}$ and $\mathbf{11.00\%}$ respectively. Furthermore, when compared to $ours$ 1.8197 and 0.7761, we observed improvements in $\mathbf{14.35\%}$ and $\mathbf{8.71\%}$. For $50 \times 50$ case, the average $RMSN$ and average $\rho$ of the $best$ method were 1.9468 and 0.9019. When compared to the $traditional$ 3.3639 and 0.6518, it showed improvements in $\mathbf{42.12\%}$ and $\mathbf{38.37\%}$ respectively. Similarly, when compared to $ours$ 3.2543 and 0.6883, we observed improvements in $\mathbf{40.17\%}$ and $\mathbf{31.03\%}$. This experimentation indicates that there is still significant room for improvement in the performance of our optimization method through further enhancements in the model and sampling method. For larger-scale matrices, this improvement is even more apparent.

\subsubsection{Ablation experiments}\label{AA}

\begin{figure}
\centering
\subfigure[Encoders]{\includegraphics[width=4.3cm]{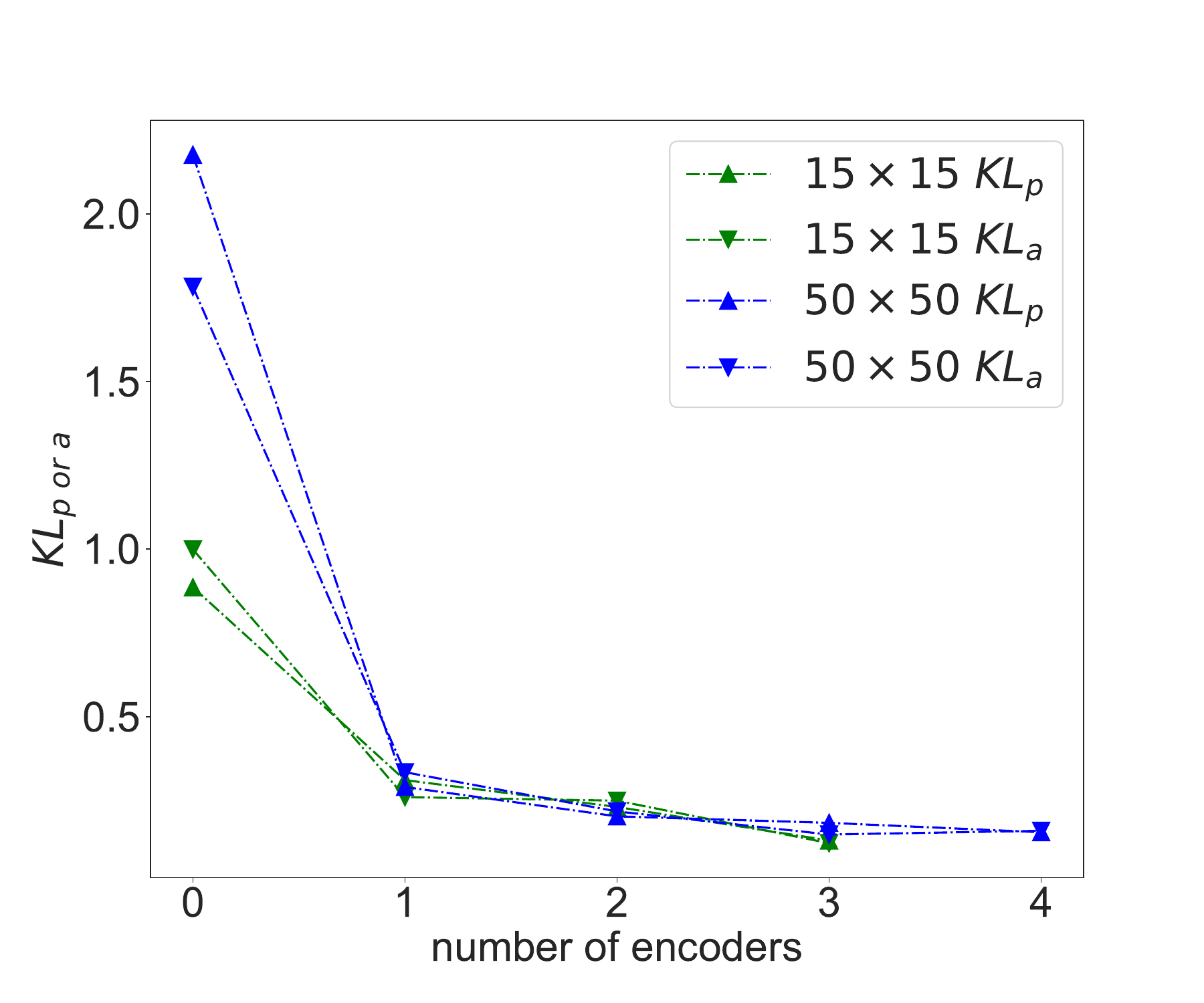}}
\subfigure[Heads]{\includegraphics[width=4.3cm]{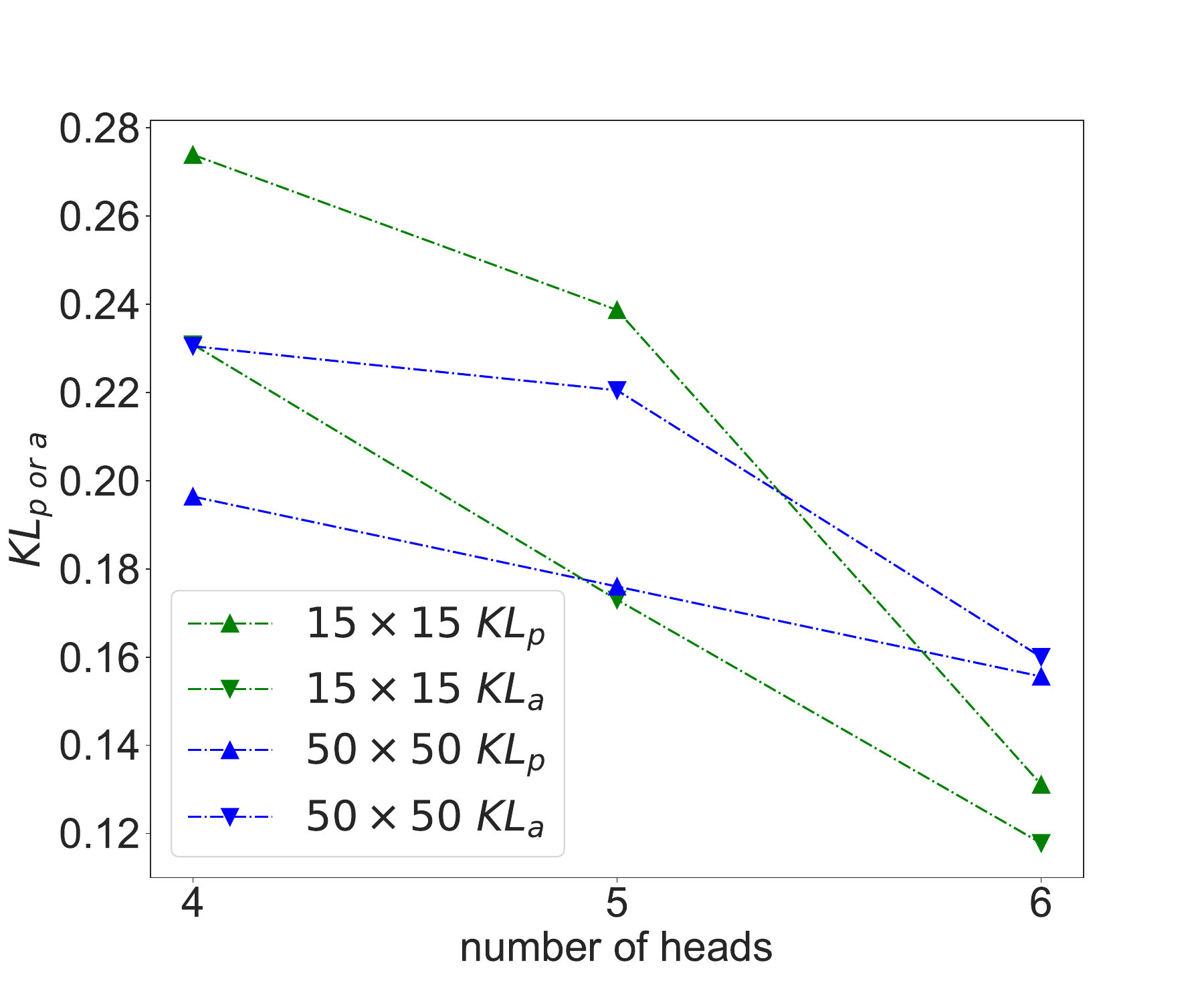}}
\caption{How the number of encoder layers and heads affect the inferred distribution $KL_{r}(\bar{\pmb{d}}_{p}||\pmb{d}_{p})$ and $KL_{r}(\bar{\pmb{d}}_{a}||\pmb{d}_{a})$}
\label{fig:2}  
\end{figure}

In our experimentation, we aimed to assess the influence of the self-attention mechanism on the model's ability to capture the pair-wise correlation of OD nodes. The findings depicted in Fig. 7(a) revealed that the model's performance enhances as the number of encoder layers increases. Moreover, the absence of a self-attention module results in a significant decline in the model's performance, highlighting the criticality of the self-attention mechanism in understanding the relationship between OD nodes.

Additionally, we explored the effect of the number of attention heads on the model's performance. The results depicted in Fig. 7(b) suggest that augmenting the number of self-attention heads provides a better comprehension of correlation features from various perspectives. However, as the number of encoder layers and attention heads increases, the computational cost also escalates, and the performance improvement saturates beyond a certain limit. Consequently, it is imperative to strike a balance between model performance and computational efficiency when choosing the number of encoder layers and attention heads in the self-attention mechanism.

 \subsubsection{Scalability of the framework with the introduction of prior matrices}\label{AA}
In Part 1, it was mentioned that a significant amount of work in OD matrix estimation is based on adding prior OD matrices as constraints on top of traffic count data, forming an important class of OD inference methods. To test the scalability of our framework, we introduced four random perturbations (p1, p2, p3 and p4) following a normal distribution to two target matrices of sizes 50x50 and 15x15, respectively. This resulted in four different prior OD matrices noted by $\pmb{t}_{prior}$ with varying degrees of bias. We used Eq. 21 as the optimization objective.

\begin{IEEEeqnarray}{c} 
\min_{\hat{\pmb{t}}}R(\hat{\pmb{t}})=\min_{\hat{\pmb{t}}}(N(\hat{\pmb{t}})+S(\hat{\pmb{t}})+prior(\hat{\pmb{t}})) \\
prior(\hat{\pmb{t}})=\frac{\Theta}{2}(\pmb{t}_{prior}-\hat{\pmb{t}})^{\mathrm{T}}(\pmb{t}_{prior}-\hat{\pmb{t}}) \IEEEnonumber
\end{IEEEeqnarray}

From Fig 7(b), (c), (d), it can be observed that incorporating the distribution information of the matrices still provides assistance in estimating the OD matrices. However, in Fig 7(a), our method did not yield better results, primarily due to the imprecise nature of the provided approximate distribution information. This can be resolved by adjusting the sampling strategy and model structure to obtain more accurate distribution information.

\begin{figure*}
\centering
\subfigure[$\rho$ of $ 50 \times 50$]{\includegraphics[width=4.4cm]{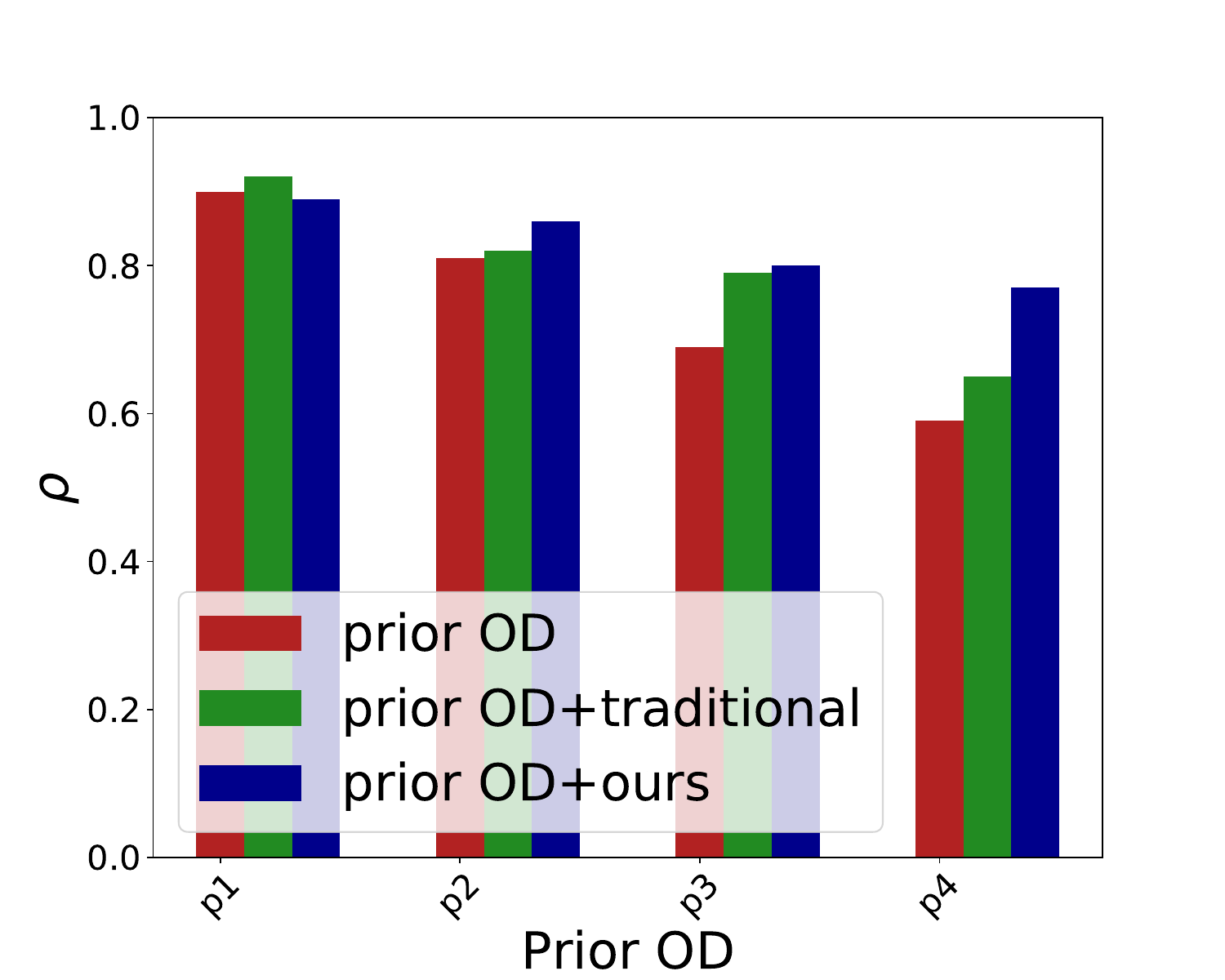}}
\subfigure[$\rho$ of $ 15 \times 15$]{\includegraphics[width=4.4cm]{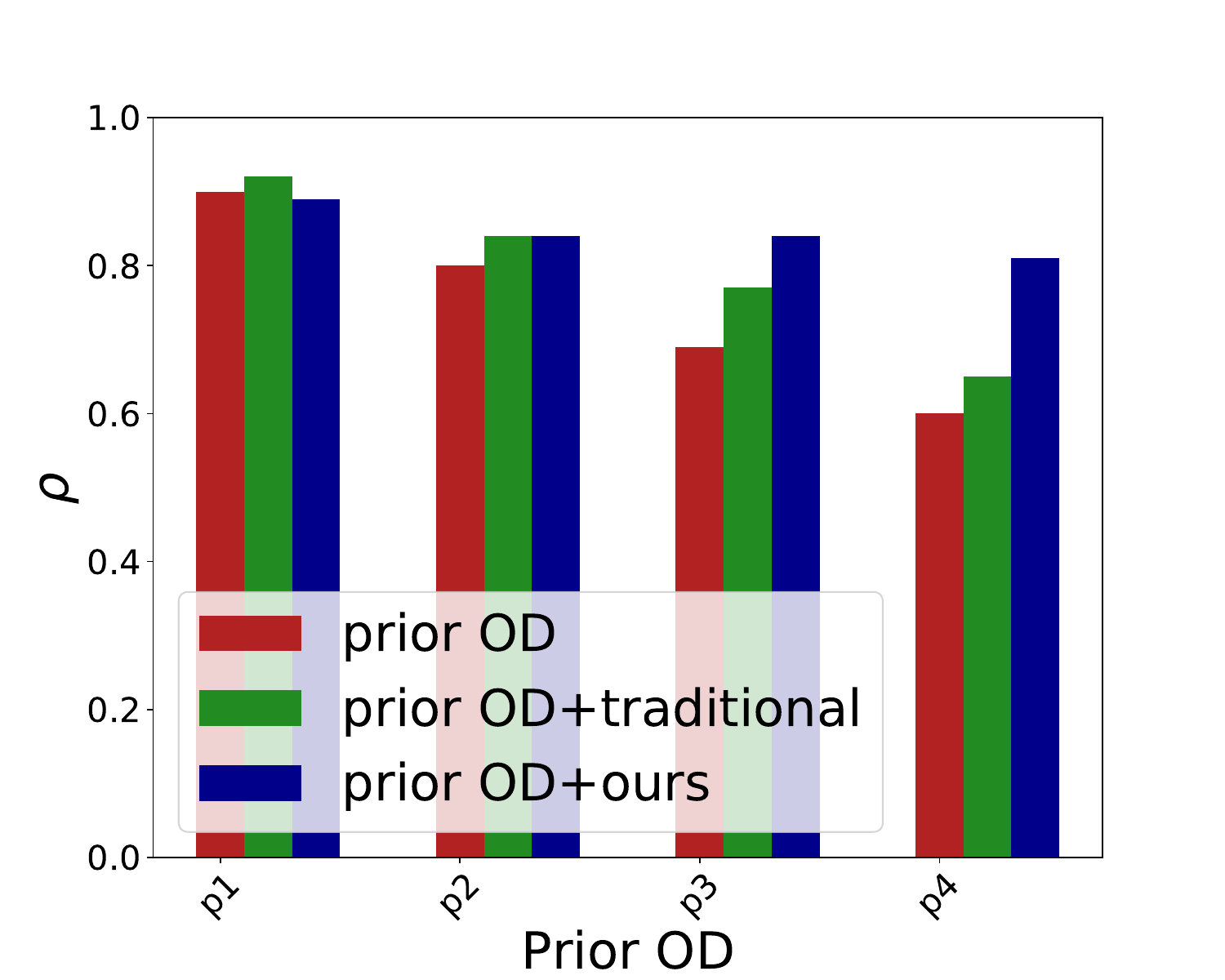}}
\subfigure[$RMSN$ of $ 50 \times 50$]{\includegraphics[width=4.4cm]{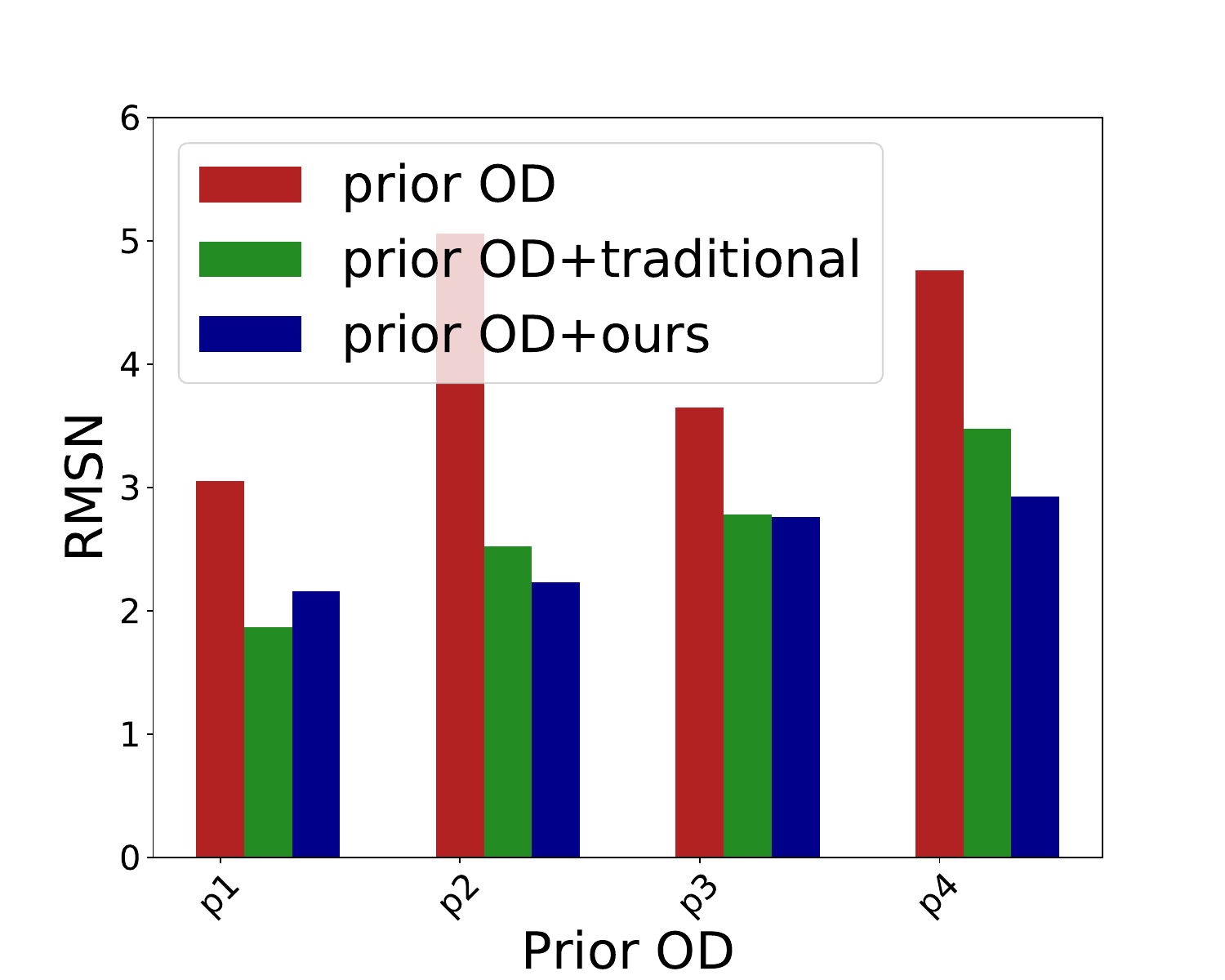}}
\subfigure[$RMSN$ of $ 15 \times 15$]{\includegraphics[width=4.4cm]{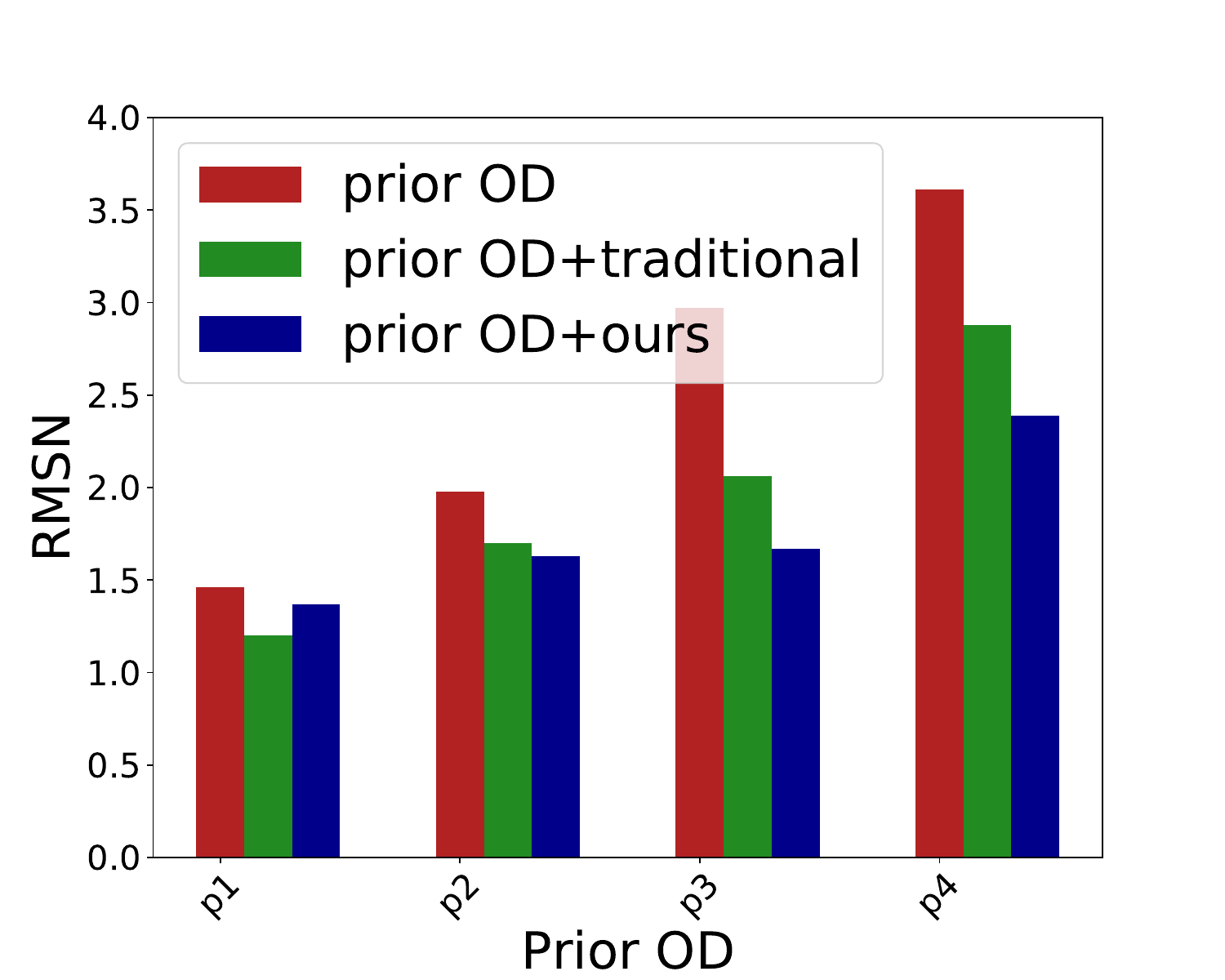}}
\caption{Comparison of optimization results between traditional and our methods after Introducing prior OD} 
\label{ }  
\end{figure*}
\subsection{Bologna}\label{AA}

\subsubsection{Study Network}\label{AA}
The dataset utilized in this study consists of actual traffic data from the Ringway Dataset for the city of Bologna\cite{bedogni2015bologna}\cite{caiati2016estimating}, located in Italy. Specifically, the data covers the morning peak hour from 8:00-9:00 o'clock. The network is divided into 15 origin-destination (OD) zones and includes 55 road sections. The target OD matrix includes a total of 22,213 individual trips, as shown in Fig. 8.

\subsubsection{Dataset}\label{AA}
In our previous experiment, we utilized randomly generated probe vehicles to create a sparse matrix, which allowed us to simulate the distribution corresponding to any possible OD matrix. While this approach provides the advantage of generalization performance for any matrix under the network, it can result in a loss of accuracy. To address this issue and better fit the NN model to the data distribution, we will utilize part of the available vehicle trajectory information as probe vehicles while considering the real OD matrix.

To generate our training dataset, we randomly and uniformly sample around 10\% trips from the independent trips and use them to form a new matrix, which can be thought of as probe vehicles with accessible GPS or trajectory data. However, this new matrix may not directly reflect the structure of the original matrix. To overcome this issue, we randomly re-sample the new matrix, avoiding OD pairs with zero trips to reduce the sampling range. By doing so, we obtain matrix data with different structures and their corresponding real traffic counts, which provide insights into actual traffic conditions, such as congestion. Our dataset consists of also around 32K sampled data pairs.

We then divide the resulting training dataset into a training set and a validation set based on an 8:2 proportion. This approach allows us to use real traffic data to improve the accuracy and fitting ability of the NN model and better reflect the actual traffic conditions under consideration.

\subsubsection{Results and analyze}\label{AA}

We conducted two experiments using the same NN model and training algorithm as in the training and inference phases, to obtain the inferred distribution $KL_{r}(\pmb{d}_{p}^{*}||\pmb{d}_{p})$=0.0122, $KL_{r}(\pmb{a}_{p}^{*}||\pmb{d}_{a})$=0.0046. During the optimization phase, initially, we set an unbiased matrix, similar to the one used in Cologne, as the starting matrix. Two different assignment algorithms were used, namely Dijikstra based on the shortest path and UE based on user equilibrium. These algorithms were implemented by "-duarouter.py" and "-marouter.py" scripts in SUMO. Our method outperformed traditional methods for both assignment algorithms, as shown in Table 4, indicating the robustness of our method to different lower-levels in the bi-level framework.

\begin{figure}
\centering
\subfigure[Bologna ringway network]{\includegraphics[width=4.3cm]{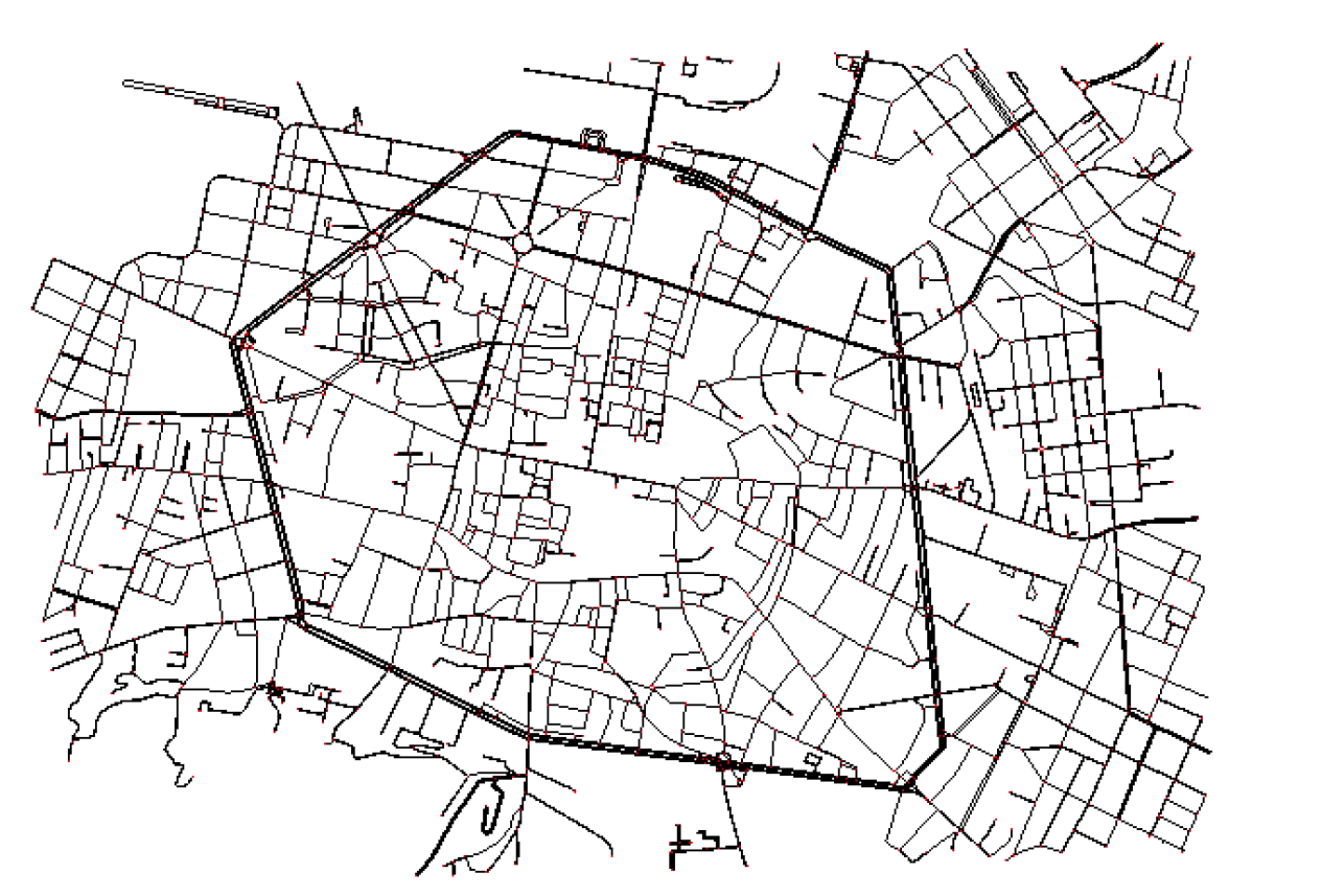}}
\subfigure[nodes and road sections]{\includegraphics[width=4.3cm]{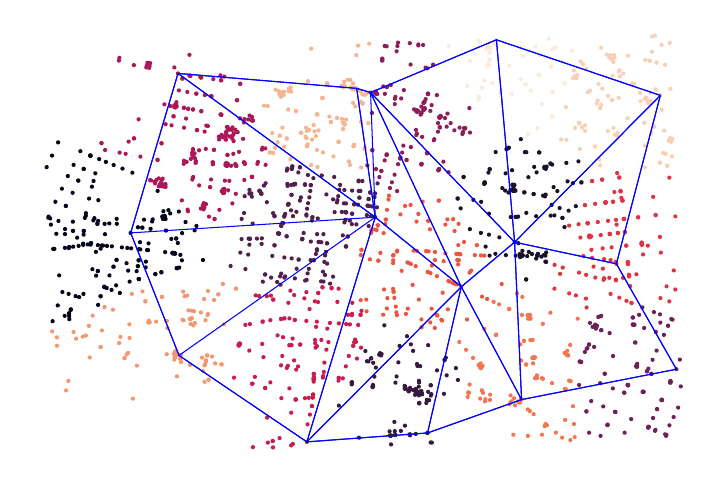}}
\caption{Real traffic data in bologna}
\label{fig:2}  
\end{figure}

Secondly, we aimed to verify the robustness of our method to different initial matrix optimizations. We obtain the second perturbation matrix by adding an error term obeying $N\left(0, 200\right)$ as the initialization matrix, and set negative values to 0. Traditional methods failed to achieve further improvement when starting from the perturbation matrix due to underdetermination, as indicated in\cite{behara2020novel}. However, our method utilized distribution information as a guide and was able to guide the optimization towards a better goal, as shown in Table 4.

\begin{table}
\centering
	\caption{}
	\label{table}
	\setlength{\tabcolsep}{3pt} 
	\renewcommand\arraystretch{1.1} 
	\begin{tabular}{m{2cm}<{\centering}|m{1.25cm}<{\centering}|m{1.25cm}<{\centering}|m{1.25cm}<{\centering}|m{1.25cm}<{\centering}}
	\toprule
		 
		Starting point& \multicolumn{2}{c|}{Initial OD 1}  & \multicolumn{2}{c}{Initial OD 2} \\ \hline
		Assignment algorithm&Dijikstra & UE  &  Dijikstra &UE    \\ \hline

		& \multicolumn{4}{c}{RMSN} \\   
		Original &\multicolumn{2}{c|}{1.4168}  & \multicolumn{2}{c}{1.8215}    \\ 
		Traditional & 1.2844 & 1.3524 & 1.6075 & 1.7450  \\ 
		Ours & \bfseries1.1464 & \bfseries1.2316 & \bfseries1.6418& \bfseries1.7081   \\ \hline

		& \multicolumn{4}{c}{$\rho$} \\   
		Original &\multicolumn{2}{c|}{0.2092}   & \multicolumn{2}{c}{0.5677}  \\ 

		Traditional & 0.4126 & 0.3805 & 0.5487 & 0.5348   \\ 

		Ours & \bfseries0.5735  &\bfseries 0.5650 & \bfseries0.6073 & \bfseries0.6026  \\ 
		\bottomrule

	\end{tabular}
	\label{tab1} 
\end{table}

 \subsubsection{Scalability of the framework with the introduction of subpath}\label{AA}
In scenarios where only a few main road segments are equipped with traffic sensors, traffic counts may only contain limited traffic information. A certain class of methods further provides structural information for OD matrix estimation by sampling Bluetooth probe vehicles to capture subpath flow patterns. Based on this approach, we tested the scalability of the framework regarding subpath flow structures.

We assumed that out of 55 road segments, only 10 (s1), 15 (s2), and 20 (s3) segments can collect traffic counts, while 22 segments can probe Bluetooth vehicles to obtain subpath structure information. As \cite{behara2020novel}  mentioned, we introduced the Pearson similarity $\rho$ of subpath $\pmb{\epsilon}_{s}$ structures as an optimization objective based on Eq. 22. From Fig 10, it can be observed that our method shows improved performance under different proportions of segments with available traffic counts.

\begin{IEEEeqnarray}{c} 
\min_{\hat{\pmb{t}}}R(\hat{\pmb{t}})=\min_{\hat{\pmb{t}}}(N(\hat{\pmb{t}})+S(\hat{\pmb{t}})+\rho(\hat{\pmb{t}})) \\
\rho(\hat{\pmb{t}})=\frac{(\pmb{\epsilon}_{s}-\pmb{\mu}_{{\epsilon}_{s}})^\mathrm{T}(\hat{\pmb{\epsilon}}_{s}-\pmb{\hat{\mu}}_{{\epsilon}_{s}})}{\sqrt{(\pmb{\epsilon}_{s}-\pmb{\mu}_{{\epsilon}_{s}})^{\mathrm{T}}(\pmb{\epsilon}_{s}-\pmb{\mu}_{{\epsilon}_{s}})} \sqrt{(\hat{\pmb{\epsilon}_{s}}-\pmb{\hat{\mu}}_{{\epsilon}_{s}})^{\mathrm{T}}(\hat{\pmb{\epsilon}}_{s}-\pmb{\hat{\mu}}_{{\epsilon}_{s}})} }    \IEEEnonumber\\
where\  
\hat{\pmb{\epsilon}}_{s}=\pmb{P}_{s}\hat{\pmb{t}} \IEEEnonumber 
 \end{IEEEeqnarray}

$\pmb{\mu}_{{\epsilon}_{s}} \in \mathbb{R}^{|\pmb{\epsilon}_{s}|}_{\geq{0}}$ is a vector with each element value equal to the mean of $\pmb{\epsilon}_{s}$, and $\pmb{\hat{\mu}}_{{\epsilon}_{s}}$ correspond to the mean-centered $\hat{\pmb{\epsilon}}_{s}$. Considering that the structural information of road segments reflects the city's topological characteristics, which is different from the provided structural information, this demonstrates the strong scalability of our framework in incorporating various constraints from different perspectives.

\begin{figure}
\centering
\subfigure[$\rho$]{\includegraphics[width=4.3cm]{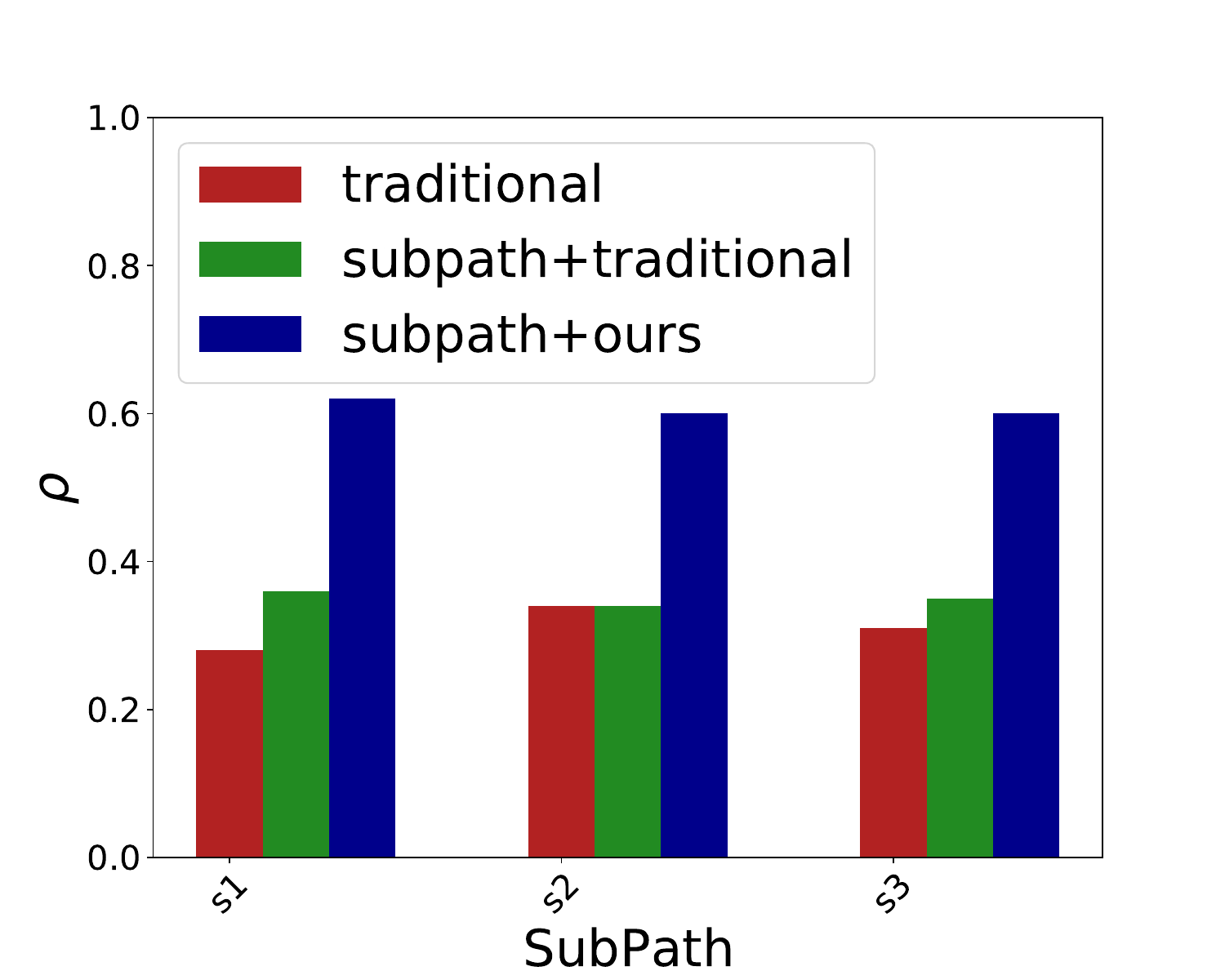}}
\subfigure[$RMSN$]{\includegraphics[width=4.3cm]{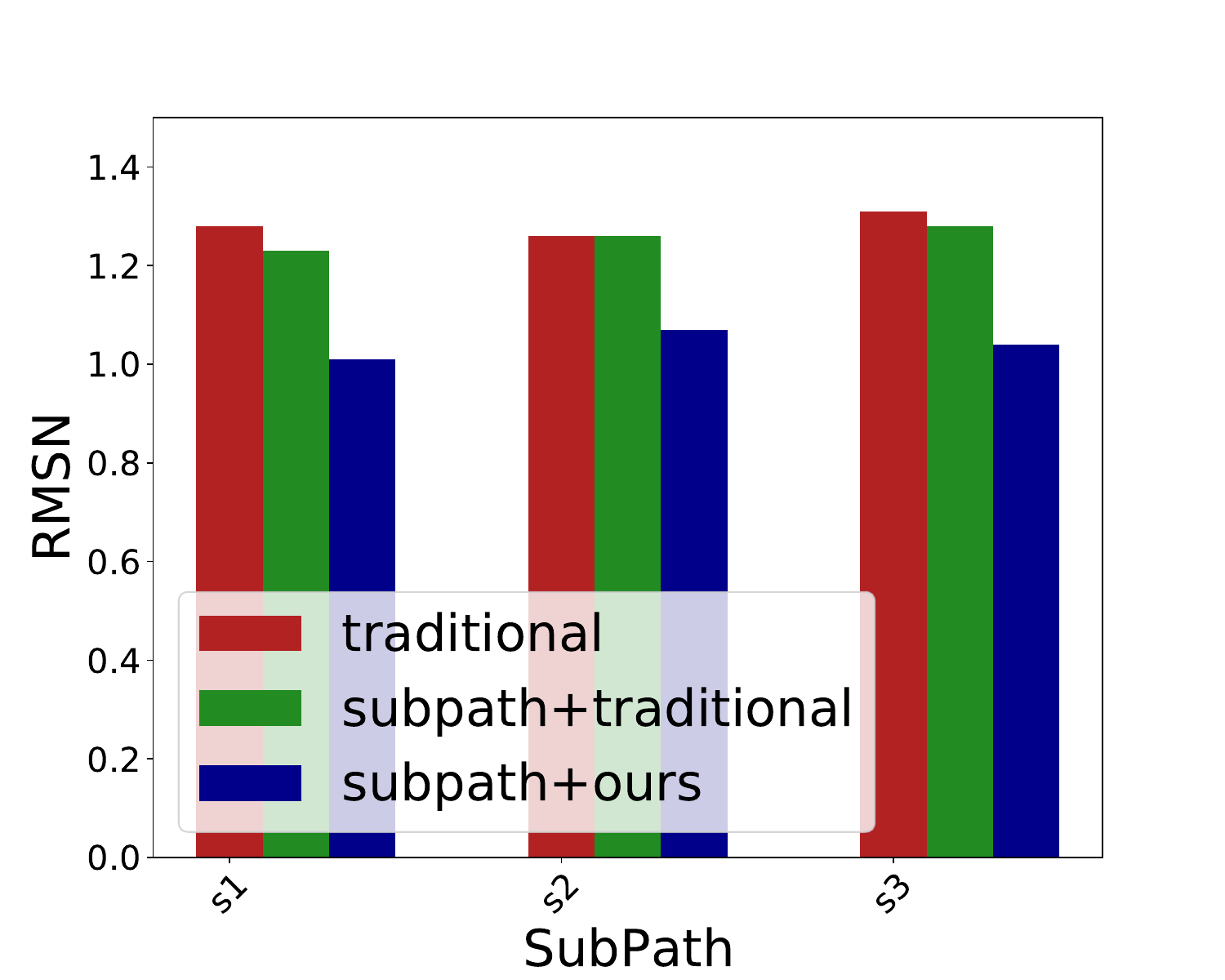}}
\caption{Comparison of optimization results between traditional and our methods after Introducing subpath}
\label{fig:2}  
\end{figure}

\section{DISCUSSION}
In summary, our method has been shown to be effective in inferring significant OD distributions by utilizing a combination of deep learning and numerical optimization algorithms. By incorporating sampling, a deep learning model, and optimization techniques, our approach can provide valuable structural information and lead to improved optimization outcomes. The success of our method can be attributed to the strengths of both deep learning and numerical optimization, which have been carefully integrated to overcome the limitations of traditional methods. Our training set is capable of quickly collecting various traffic distributions across the entire city, despite being limited to sampling from sparse matrices and a small set of values. In our experiments, we were able to gather 32K training data in just 24 hours using ten cores at 2.40 GHz. Overall, our approach offers a promising new direction for improving the accuracy and efficiency of traffic distribution inference, with potential applications in transportation planning and management.

We are putting forward a simple deep learning model that solely takes into account topology and traffic data. Our belief is that by integrating additional features and enhancing the model architecture, we can derive more precise distribution information, thereby enhancing the optimization outcome, which will be our future works.


\section{CONCLUTION}
In this paper, we introduce a deep learning model that maps traffic count distributions to OD distributions while taking into account the city network's topological structure and the pairwise dependence between OD nodes.  Furthermore, we extend the numerical optimization of the bi-level framework by incorporating the optimal inferred OD distribution. We demonstrate the feasibility of our approach on a real large-scale city network and a real traffic dataset. Our experiments indicate that our method is effective and provides both numerical and structural improvements to the estimated OD matrix compared to traditional methods.

However, given the NN's powerful capabilities and high flexibility, we aim to expand the NN's structure and sampling method to enhance inference and optimization performance. Additionally, we intend to apply our method to a larger volume of real traffic data. Specifically, we aim to analyze datasets that contain genuine GPS data which could be considered as probe vehicles. This represents a key focus area for our future work.
\bibliography{reference}

\bibliographystyle{IEEEtran}

\end{document}